\documentclass[review,3p]{elsarticle}

\usepackage{lineno,hyperref}
\modulolinenumbers[5]

\journal{European Journal of Operational Research}

\biboptions{authoryear}

\usepackage{booktabs} 
\usepackage{setspace}
\usepackage{enumerate}
\usepackage{amssymb,mathtools} 
\usepackage{algorithm}
\usepackage{algorithmicx}
\usepackage[]{algpseudocode}
\usepackage{amsmath}
\usepackage{amsthm}
\usepackage{caption}
\usepackage{array} 
\usepackage[dvipsnames,table]{xcolor}
\usepackage{soul}

\usepackage{subcaption}

\newtheorem{proposition}{Proposition}

\begin{document}

\begin{frontmatter}

\title{Scheduling and Routing with Degradation-Triggered Job Arrivals: An Application to Forest Firefighting with an Unmanned Aerial Vehicle Fleet}

\author[Hacettepe]{Erdi Dasdemir}
\ead{edasdemir@hacettepe.edu.tr}

\author[Buffalo]{Esther Jose}
\ead{estherjo@iastate.edu}
 
\author[Buffalo]{Rajan Batta\corref{mycorrespondingauthor}}
\cortext[mycorrespondingauthor]{Corresponding author}
\ead{batta@buffalo.edu}

\address[Hacettepe]{Department of Industrial Engineering, Hacettepe University, 06800 Ankara, Turkey}

\address[Buffalo]{Department of Industrial and Systems Engineering, University at Buffalo (SUNY), Buffalo, NY 14260}






\begin{abstract}
We define an intertwined scheduling and routing problem where new jobs appear due to the degradation of the existing jobs. Specifically, once a job arrives at a potential job location, a time window begins during which the demand of the job can be fulfilled. The demand degrades within the time window, and once it surpasses a particular threshold, it triggers the arrival of new jobs. Each job location inherently possesses an initial default reward, and the presence of an unprocessed job at a location gradually reduces this default value. The overall objective is to maximize the total remaining reward. The underlying motivation of this problem aligns with the proverb ``a stitch in time saves nine," and the problem itself carries practical implications. We focus on the problem in the context of aerial forest firefighting. Each ignited area has a designated action window; delaying intervention causes the fire to grow, diminishing the area's value and causing it to spread to adjacent areas. We develop a mixed-integer programming model that maximizes value retention in wildfire-threatened regions, and a hybrid model based on dynamic constraint generation to enhance the scalability of the model. We evaluate the performance and practicality of our models through computational experiments and a case study. Additionally, we ensure the study's reproducibility and encourage further research by providing open access to the codebase of our model.

\end{abstract}


\end{frontmatter}

\newpageafter{abstract}

\pagebreak

\section{Introduction}
Scheduling and routing problems hold significant importance in the realm of operations research (OR) due to their broad applicability in real-world contexts. Both of these problems typically aim to align the available workforce with the given workload. The scheduling problem places a specific emphasis on temporal considerations such as sequence and timing of service deliveries, whereas the routing problem focuses on spatial factors like job locations and travel expenses. In many applications, the temporal aspects of scheduling and the spatial aspects of routing are intertwined, where the timing of service matters in routing, and travel duration holds significance in scheduling. In these problems, typically, demand is assumed to be known before the execution phase. However, in certain real-world scenarios, new jobs may reveal themselves during the execution phase. While this can sometimes be attributed to random factors, it might also be determined by decisions taken and the inherent attributes of problem variables. In this study, we address a novel routing and scheduling problem where new jobs emerge during the execution phase, and their arrivals are triggered by the degradation of existing jobs.

We consider a network comprising a set of potential job locations, each with corresponding rewards, and a set of initial jobs located in some of the locations. The objective is to identify the optimal scheduling and routing decisions for a fleet of mobile servers, to maximize the total collected reward across the entire network. Each potential job location within the network could serve as a destination for arriving jobs, and the default reward of the location is maintained when it remains job-free. A job arrives with a specific time window during which its demand can be fulfilled. The demand for a job degrades within its designated time window, and once it surpasses a particular threshold, it triggers the arrival of new jobs. In addition, the degradation of a job gradually reduces the default reward of its location. Consequently, the job arrivals in the network are dynamic in terms of quantity, location, and time. Rather than being random, this dynamic nature is determined by problem variables and scheduling and routing decisions. Then, there is a two-way interaction---decisions impact new arrivals, and new arrivals impact decisions. We classify this problem as a variant of the Vehicle Routing Problem (VRP) \citep{Braekers_2016} and Job Scheduling Problem (JSP) \citep{Zhang_2019} that incorporates rewards, time windows, and degradation-triggered new job arrivals.

We study this problem in the context of aerial forest firefighting using a fleet of unmanned aerial vehicles (UAVs). Given a forested region prone to wildfires, the objective is to protect the value of the region by suppressing any existing fires and preventing further fire spread. Each ignited spot is considered a job. Retrieving water from water sources and delivering it to extinguish the fire at an ignited spot is considered to be a service. When a spot catches fire, a time window is initiated to address the fire. As time elapses, the fire gradually increases in size, resulting in a reduction of the value of the spot and its surroundings. Moreover, if a spot remains unprocessed until its size reaches a certain threshold, the fire spreads to neighboring locations, giving rise to new jobs that require processing. We utilize UAVs in this study to motivate ongoing initiatives on their usage in aerial firefighting but there is no practical difference in the execution of the service regardless of whether the vehicles are manned or unmanned. This problem echoes the adage ``a stitch in time saves nine” and addresses a relevant problem in which early and well-coordinated intervention can prevent disproportionate damage.

We develop a mixed integer programming (MIP) model that aims to find the optimal scheduling and routing decisions that maximize the total remaining value of the entire region. Our model addresses several aspects of decision-making for the problem. Given initial ignitions at fire-prone nodes, the model aims to schedule and route a fleet of firefighting UAVs to maximize the total preserved value. Routing and scheduling decisions include job selection, vehicle-job assignments, visit order, water refill locations, and arrival times. The model also determines fire spread across the network, including the timing and location of new fires, whether a node is processed, and the remaining value at each node. To enhance the scalability of the MIP model for difficult instances, we develop a hybrid model that builds on a relaxed formulation and leverages a branch-and-cut framework with dynamic constraint generation. We conduct computational experiments to evaluate the performance of the model and explore the effects of problem parameters on solution performance. 
To further illustrate its practical usage, we develop a case study grounded in the actual characteristics of Californian wildfires. We showcase our model across various scenarios with different regional characteristics. We also openly share our codebase on GitHub\footnote{\href{https://github.com/edasdemirlab/job-scheduling-fire-2024.git}{https://github.com/edasdemirlab/job-scheduling-fire-2024.git}: The repo is private to keep the code confidential before publication. We have included our codebase as a supplementary file to facilitate its review by the referees.} to promote reproducibility and assist practitioners in replicating our results and building upon our work in the future. Users can engage with our codebase, which functions as a software with a spreadsheet-based interface. They can create new problem instances or seek optimal solutions for their specific instances simply by preparing the necessary inputs via the spreadsheet.

To summarize, our contributions are threefold:
\begin{itemize}
    \item We contribute to the scheduling and routing literature by considering degradation-triggered job arrivals. We develop a MIP model to identify optimal scheduling and routing decisions in this context\textcolor{NavyBlue}{, and utilize it within a hybrid model that leverages dynamic branch-and-bound and cut generation to improve performance.} 
    \item We contribute to the field of disaster management by addressing integrated logistics and operational planning challenges associated with forest fire management. We develop a prescriptive decision-making model designed to provide recommendations to decision-makers involved in responding to forest fires using aerial vehicles. 
    \item We develop and share our codebase, which functions as a software with a spreadsheet-based interface. The software can be used to replicate our results, generate new problem instances, or find optimal solutions given problem instances.

\end{itemize}

The paper is organized as follows. In Section 2, we review the relevant literature on scheduling, routing, and forest firefighting, highlighting our contributions. In Section 3, we define the problem statement and present a demonstrative toy example. In Section 4, we develop our mathematical model and conduct and validate its framework. In Section 5, we present computational experiments to evaluate the performance and limitations of the model. 
In Section 6, we introduce a case study based in California to demonstrate the practical applications of our model. Finally, in Section 7, we conclude the paper by summarizing our key findings and contributions, and by outlining avenues for future research.

\section{Relevant literature}
We start with providing a concise overview of scheduling and routing problems to establish the required foundation for our problem. Following that, we provide a review of the relevant works in the field of forest fire fighting. During the review, we highlight the distinctive characteristics of our study, position it within the existing landscape of similar problems in the literature, and specify the gaps our study addresses.

\subsection{Scheduling and routing}

The prominent example of scheduling problems is the Job Shop Scheduling Problem or the Job-Shop Problem (JSP) (see \citealp{Zhang_2019}, for a detailed review). An important extension closely related to our study is Job Scheduling with Degradation, where the workload requirements of jobs increase over time (see \citealp{Aarabi_2020}, for an example). The fundamental example of routing problems is the Vehicle Routing Problem (VRP) (see \citealp{Braekers_2016}, for a comprehensive review). When executing a job yields a benefit, as in our case, the VRP becomes the VRP with Profits (VRPP) (see \citealp{Stavropoulou_2019}, for an example). 

In many real-world applications, the temporal aspects of scheduling and the spatial aspects of routing are intertwined and the problem becomes a mix of vehicle routing and scheduling problems. Several naming conventions have been employed in the literature to describe the mixed problem. Examples include Vehicle Routing-Scheduling (c.f. \citealp{Huang_2015}) and Scheduling and Routing Problem (c.f. \citealp{Guastaroba_2021}). The literature also includes extensions of the main problems, such as Job Scheduling with Travel Times (c.f. \citealp{Mejia_2020}) or VRP with Time Windows (VRPTW) (c.f. \citealp{Quirion-Blais_2021}. In a typical VRPTW, a fleet of mobile servers is scheduled to process a set of spatially distributed jobs, where jobs need to be processed in their specified time windows. \textcolor{NavyBlue}{\citet{BERGHMAN20231} emphasize in their review that integrating production scheduling with outbound vehicle routing yields significant performance improvements and highlights the growing interest in jointly addressing these interconnected problems.}

Our problem involves both scheduling and routing decisions, making it an extension of the literature on both problem types. The degradation and time window aspects of our problem are associated with Job Scheduling with Degradation and VRPTW extensions. Specifically, we incorporate both soft and hard time windows. The service interval at a fire location is hard because the vehicles can respond to the fire only after it starts and before it extinguishes by itself (a fire extinguishes by itself once everything at the location is burned out and nothing remains to keep the fire ongoing). However, the time windows can also be considered as soft as UAVs can serve fires any time before the fire extinguishes by itself but late arrivals incur penalties in the form of a lower remaining value at the job location and creation of new jobs at the nearby locations.

To clarify, the concept of degradation in our study diverges from the conventional approach in job scheduling. In standard practice, degradation is understood to increase the processing requirements of a job. As an example of the traditional approach, we refer to \citet{Aarabi_2020}. In our consideration, a degraded job can continue to be fulfilled based on its original demand but gives rise to new job arrivals. The new jobs have their own distinct requirements, and whether they will receive services is a new decision task. 

We also identify our problem as closely related to the Dynamic VRP, an extension of VRP where inputs dynamically reveal themselves during the execution of routes (see \citealp{PILLAC20131}, for a review on Dynamic VRP). However, existing literature often considers demand arrivals that are independent of routing and scheduling decisions. The typical assumptions include known appearance times or arrival probability distributions. For instance, \citet{dearmasetal2015} and \citet{GHANNADPOUR2014504} consider known arrival times for new requests, and \citet{ALBAREDASAMBOLA201431} consider known probability distributions that model the set of customers requiring service in later time periods. In these studies, there is a one-way interaction where dynamic arrivals influence routing and scheduling decisions. In contrast, in our problem, the interaction between new job arrivals and routing and scheduling decisions is two-way. This implies that routing and scheduling decisions significantly impact the determination of job arrivals, and in turn, job arrivals have a substantial influence on routing and scheduling decisions.

\subsection{Related work in forest firefighting}

Forestry and OR share a long-standing relationship, with the former presenting various potential OR problems and the latter providing optimal recommendations for these problems. \citet{Ronnqvistetal2015} provide an overview of the current status of OR applications in forestry and identify 33 open problems. Our research aligns with these open problems, specifically addressing the ones related to fire management. We address the deployment of air vehicles for long-duration suppression operations and the development of a fire management decision support system for effectively managing extensive fires.

\textcolor{NavyBlue}{Modeling fire behavior in continuous time is a major challenge in forest firefighting and is a complex research area with numerous models available \citep{CRUZ201316}. Optimization studies in firefighting often avoid this complexity by assuming discrete-time updates for fire conditions (see, for example, \citealp{ZHOU2019106101} and \citealp{AVCI2024488} for recent studies). In continuous time, although non-linear functions may represent fire behavior more accurately, they are difficult to integrate into optimization models. In our approach, we strike a balance between realism and tractability by modeling fire spread and value degradation in continuous time using a linear approximation.}

In forest fires, `spotfires' are a major reason for wildfire spread, as discussed by \citet{storey2020drivers, storey2021experiments} and \citet{cruz2012anatomy}. Unlike the typically modeled spread of wildfire through direct contact with vegetation or fuel, spotfires are small fires that ignite near existing wildfires due to pieces of flaming material called firebrands that jump from existing areas on fire to nearby areas. In our work, while the degradation of a job within a node represents fire spread via contact, the creation of new jobs or ignitions nearby represents fire spread via spotfires.

While our primary focus lies in optimizing aerial firefighting through a routing and scheduling perspective, we also contribute to the forest fire management literature by accounting for new fire arrivals resulting from the degradation and spread of existing ones. In terms of considering the degradation in existing fires, three studies in the literature similar to ours are \citet{Rachaniotisetal2006}, \citet{PappisAndRachaniotis2009}, and \citet{PappisAndRachaniotis2010}.

\citet{Rachaniotisetal2006} consider traditional degradation in fire tasks, where delays in fire response result in value loss and increased time for suppression. Yet, no new fire arrivals are considered. Their work focuses on a single firefighting resource responding only to known fire tasks. The authors concentrate on modeling the rate of spread to determine the required suppression time rather than utilizing optimization approaches, providing only a small example with four fires where all possible schedules can be enumerated. \citet{PappisAndRachaniotis2009} extend this work by incorporating setup times that consider the travel time of a ground processor from one location to another. \citet{PappisAndRachaniotis2010} also expand on the work of \citet{Rachaniotisetal2006} by considering multiple vehicles and developing a heuristic approach. The key difference in our approach is that the degraded fires in our case result in fire spread and new fire tasks. Their assumption about fire spread is valid only within the job location, and spread to adjacent locations is not considered.

Another similar study to ours is \citet{HongGuang2020}. They explore a dynamic optimization problem using a simulation-based solution approach. Their model involves a probabilistic state transition structure. In comparison to our approach, they do not consider degradation in fire size and region value; instead, their focus is on minimizing the forest fire rescue time.

Regarding the value of the region, in forest fire management, distinct areas within a region may be assigned different priorities based on factors such as proximity to urban areas or the presence of critical assets like electrical substations and bridges. The task of determining the optimal allocation of resources to minimize loss is commonly known as the Asset Protection Problem (APP), \textcolor{NavyBlue}{a recent variant of the Team Orienteering Problem (TOP) introduced by \citet{vander2015}). To solve this problem more effectively, \citet{Yahiaoui2023} develop a hybrid heuristic combined with a set covering-based post-optimization and demonstrate significant improvements on benchmark instances. Another related idea to value loss can be found in the Hazardous Orienteering Problem (HOP) introduced by \citet{Alberto2023}, where visiting a location earns profit, but if a time-based catastrophic event occurs, all profit is lost.}

For a more in-depth exploration of the APP and its application in forest fire fighting to optimize resource deployment plans, see \citet{Roozbeh2021}. In line with widely accepted practices, our study assumes that different sections within a region possess varied values. The emphasis on minimizing the loss of value in our problem establishes a connection to the APP.

\textcolor{NavyBlue}{Our problem also relates to dynamic scheduling and routing in disaster response, (see \citealp{WOHLGEMUTH2012261}, \citealp{WEX2014697} and \citealp{MAYADUQUE2016272} for some examples), but key distinctions arise. Once ignited, forest fires evolve continuously, driven by both natural dynamics and firefighting actions. This results in complex spatial-temporal interactions that require anticipatory planning. Refilling needs in aerial operations add further layers to the decision-making process.}

\section {Problem statement}
We consider a forested region characterized by distinct components: fire-prone (ignitable) areas, fire-proof (fire-resistant) areas, water sources, and a base for UAVs. We map this region using a grid framework, where each grid cell represents one of these components. We adopt the term ``node" to refer to each grid cell. The problem can then be represented by graph $G=(N,A)$, where $N$ encompasses all nodes, an arc set $A=\{(i,j) \mid i,j \in N, \ i \neq j\}$, where each arc is a direct link between two nodes, and a fleet of vehicles represented by the set $K=\{1,2,...,k\}$. 

A fire can only arrive at a fire-prone node without fire, and this can be triggered only by neighboring fire-prone nodes having active fire. Initially, we assume that active fires at some nodes are known. If a fire is serviced by a UAV before the end of its time window, the node is rescued; otherwise, it burns down. We illustrate the possible states of nodes with a state transition diagram in Section A of Supplementary Material.

\begin{figure}[ht] 
\centering
\includegraphics[scale=0.8]{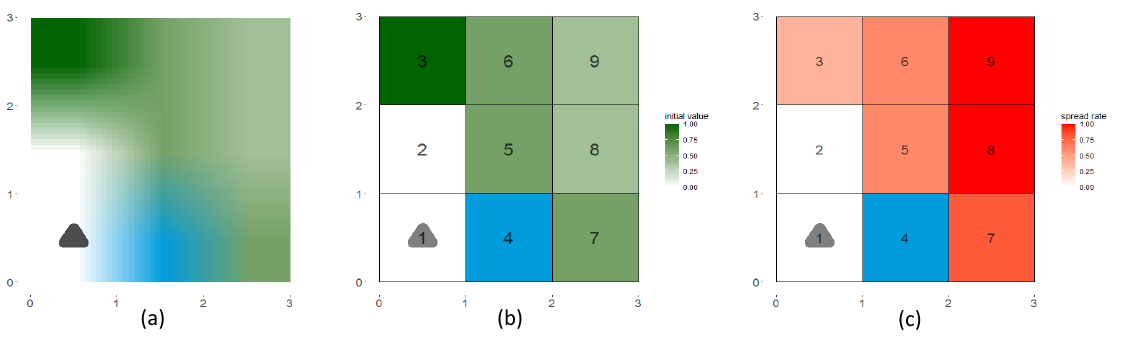}
\caption{Demonstration of problem structure}
\label{figure:problem-definition-demonstration}
\end{figure}

Each fire-prone node possesses two regional characteristics: an initial value representing its significance (reward) for decision-makers, and a spread rate indicating its potential for fire propagation. We provide an illustrative example in Figure \ref{figure:problem-definition-demonstration}. Figure \ref{figure:problem-definition-demonstration}(a) shows a forested region spanning a 3x3 area. At this stage, we can consider the dimensions of the region to be unitless. We present two grid sketches for this region. Figure \ref{figure:problem-definition-demonstration}(b) represents the default rewards of the nodes, while Figure \ref{figure:problem-definition-demonstration}(c) illustrates the spread rates of the nodes. In both figures, the darker colors indicate the higher values. Among the nodes, 1 and 2 are fire-proof nodes, 4 serves as a water resource, and the remaining are fire-prone nodes. The base is situated at node 1, serving as the point where UAVs start and complete their routes. Given some initial ignitions at some fire-prone nodes, the objective is to determine the optimal execution plan for a fleet of firefighting UAVs to maximize the total value maintained from the entire graph. 

\subsection{Degradation in fire size and node value}

\begin{figure}[ht] 
\centering
\includegraphics[scale=0.8]{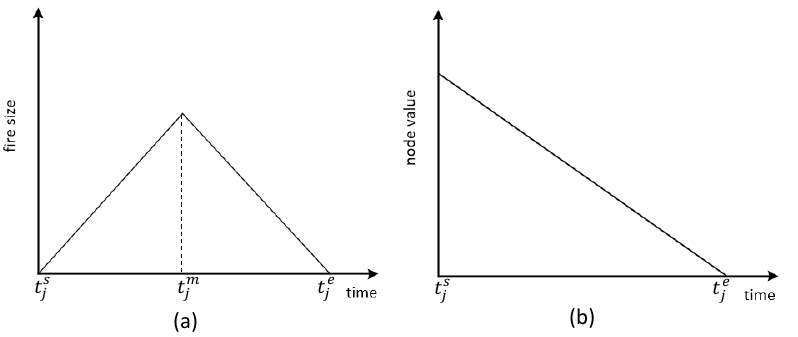}
\caption{The demonstration of degradation in fire size and node value}
\label{figure:problem-definition-time-window}
\end{figure}

Initially, time is set to zero, $t=0$. At $t=0$, the information about fire-prone nodes with active fires, represented by subset $N_a$, is known. When a fire arrives at node $j \in N_f$, it comes with a service time window starting from the arrival time $t_j^s$ and ending at $t_j^e$ when everything is burned down. Within this time window, node $j$ degrades in both its fire size and default value. \textcolor{NavyBlue}{We use linear approximations for both types of degradation, as illustrated in Figure \ref{figure:problem-definition-time-window}, to balance practicality and tractability. Incorporating non-linear functions would increase complexity and shift focus away from operational decision-making. \citet{CRUZ201316} show that fire behavior modeling involves high uncertainty, with substantial variation between predicted and observed outcomes. This supports our choice to prioritize optimizing firefighting operations over modeling detailed fire dynamics.} We incorporate factors like soil structure, moisture, and wind via the node's spread rate. We refer to \citet{dasdemir_ismet_2020} for practical forest fire behavior factors.

Figure \ref{figure:problem-definition-time-window}(a) illustrates fire size degradation. The fire originates from an infinitesimally small spot at start time $t_j^s$ and spread across the grid by $t_j^m$ based on the spread rate of node $j$. If not addressed by a UAV by $t_j^m$, it spreads to adjacent nodes in node set $j^+$. Adjacent nodes share a grid edge with node $j$. After $t_j^m$, the fire size decreases as the burned area cools, and the entire grid is burned down by $t_j^e$. 

\textcolor{NavyBlue}{The second aspect of degradation concerns node value, illustrated in Figure~\ref{figure:problem-definition-time-window}(b). In the absence of fire, each node retains its full default value, which is collected without requiring intervention.  Upon fire arrival at $t_j^s$, the node's value begins to decline until it is either suppressed or fully lost at $t_j^e$, thereby encouraging timely suppression rather than unnecessary fire creation.}

\subsection{Aerial fire processing by UAVs}
We utilize UAVs to contribute to the ongoing initiatives aimed at incorporating them into aerial firefighting operations (see, for example, \citealp{Kumar_Cohen_2011}, \citealp{Yuan2015}, and the K-Max unmanned helicopter on platforms like YouTube). However, the execution of the firefighting service in our study is unaffected by whether the vehicles are operated by humans or are unmanned. 

A fire-processing UAV addresses an active fire by releasing a load of water, with the assumption that this process is instantaneous and effectively puts out the fire within the node. \textcolor{NavyBlue}{All node (grid) areas are uniform in size, corresponding to the maximum area a UAV can extinguish with its water capacity. The UAV always releases its full water load, regardless of the fire size within the node.} The set denoted as $N_w \in N$ encompasses the water resources throughout the region, which the UAVs must access to refill their tanks between successive fire-processing tasks.

\subsection{Demonstration on a toy instance}
\begin{figure}[h] 
\centering
\includegraphics[scale=0.9]{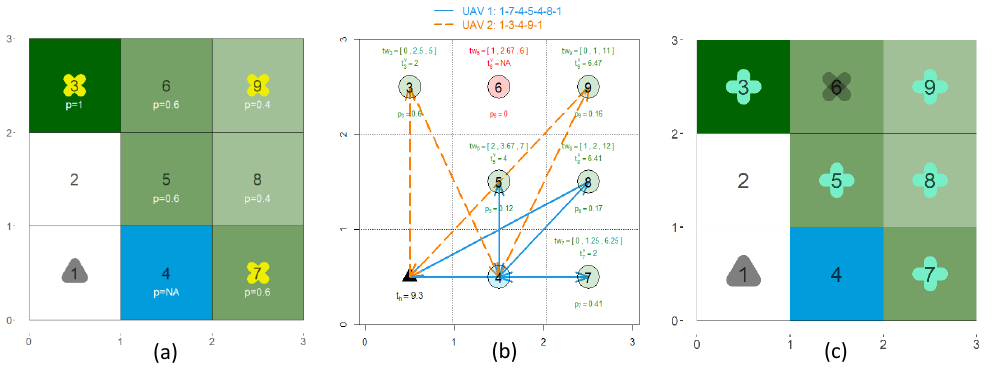}
\caption{Demonstration of a firefighting plan}
\label{figure:demonstration_solution}
\end{figure}

Figure \ref{figure:demonstration_solution}(a) shows an example region with 1 base (node 1), 1 fire-proof grid (node 2), 1 water resource (node 4), and 6 fire-prone nodes (nodes 3, 5, 6, 7, 8, 9), totaling 9 nodes. Active fires are initially at nodes 3, 7, and 9, marked by yellow ``x" symbols. Each grid display its current value ($p$), arbitrarily set between 0 and 1 for this example. There are 2 UAVs, each moving at a speed of 1 unit of distance per unit of time.

Figure \ref{figure:demonstration_solution}(b) portrays the optimal solution to the problem. UAV 1 departs from base 1 and processes the fires at nodes 7, 5, and 8. During its movement between these nodes, it stops at water resource 4 for refilling. On the other hand, UAV 2 is allocated to nodes 3 and 9, pausing only once at the water resource 4 between the two nodes. In the plot, each node $j$ is associated with a time window, $tw_j = [t_j^s, t_j^m, t_j^e]$, and the time of the service provided by UAVs, $t_j^v$, Notably, while Figure \ref{figure:demonstration_solution}(a) displays three initial fires, Figure \ref{figure:demonstration_solution}(b) depicts six fires. This indicates that new fire arrivals are triggered by the existing ones. Specifically, fire spreads from node 9 to nodes 6 and 8, as the UAV's arrival time at node 9, $t_9^v=6.47$, surpasses $t_9^m=1$. As a result, new time windows are initiated for the fires at nodes 6 and 8, starting at time 1. The fire at node 8 triggers another fire at node 5 at time $t_8^m=2$, as the UAV arrives at at $t_8^v=6.41$. Importantly, out of these six fires, five are managed by the two UAVs, maintaining their corresponding remaining values. However, the fire at node 6 is left to its natural course and eventually extinguishes itself at $t_6^e=6$, leading to a complete loss of value at the node ($p_6=0$). We see the resulting situation at the region in \ref{figure:demonstration_solution}(c), where nodes 3, 5, 7, 8 and 9 are rescued while node 6 is burned down.

\section{Solving the problem using mathematical optimization}
In this section, we develop an MIP model and examine its properties. \textcolor{NavyBlue}{We then introduce a hybrid model that leverages the MIP model to guide and structure its solution process.} To start, we present the notation used for the mathematical optimization in Table \ref{table:notation}.

\renewcommand{\arraystretch}{0.55} 
\begin{table}[ht!]
\caption{Mathematical Notation}
\centering
\begin{tabular}{r  p{11cm}} 
\toprule
\underline{Sets} &    \\ 
$N$ & Set of all nodes \\
$N_a$ & Set of the nodes with active fires at the start. \\
$N_w$ & Set of the nodes that are water resources. \\
$N_p$ & Set of the nodes that are fire-proof. \\
$N_f$ & Set of the nodes that are not fire-proof, i.e. $N_f = N {\setminus} \{N_w {\cup} N_p {\cup} {h}\}$. \\
$K$ & Set of vehicles. \\
$j^+$ & Set of nodes neighboring $j$. \\
\underline{Parameters} &    \\ 
$n$ & Number of nodes. \\
$h$ & Home base for UAVs. \\
$\delta_j$ & Region area of node $j$.  \\
$\tau_j$  & Initial value of region $j$. \\
$\beta_j$ & Value degradation rate at node $j$. \\
$d_j$ & Fire degradation rate at node $j$. \\
$\alpha_j$ &  Fire amelioration rate at node $j$.  \\
$a_j \in \{0,1\}$ & Binary parameter indicating if node $j$ has an active fire at the start. \\ 
$d_{i,j}$ & Travel duration from node $i$ to $j$. \\
$\lambda$ & Sufficiently small penalty coefficient for the time to return to base.\\ 
$\varepsilon$ & Sufficiently small positive constant.\\ 
$t_{max}$ & Maximum time allowed for the scheduled operation.\\
\underline{Decision Variables}&    \\ 
$x_{i,j}^k \in \{0,1\}$ & Binary variable indicating if vehicle $k$ moves from node $i$ to $j$. \\ 
$w_{i,j,l}^{k} \in \{0,1\}$ & Binary variable indicating if the water resource $l$ is used for refilling by vehicle $k$, when moving from node $i$ to $j$. \\
$z_{i,j} \in \{0,1\}$ & Binary variable indicating if the required conditions for a fire in node $i$ to spread to node $j$ have been met. \\
$q_{i,j} \in  \{0,1\}$ & Binary variable indicating if a fire arriving at node $j$ was triggered by a fire at node $i$.\\
$y_j \in \{0,1\}$ & Binary variable indicating if there is a fire arrival at node $j$ during the scheduled operation. \\
$s_j^c \in \{0,1\}$ & Binary auxiliary variables that are used to determine the fire spread at node $j$, where $c \in {1,2,3,4}$ corresponds to the four spread cases explained in Section 4.4.\\
$b_j \in \{0,1\}$ & Binary variable that equals 1 if a fire arrives at node $j$ but remains unprocessed, leading to the region burning down naturally, and 0 otherwise, indicating either a processed fire or no fire arrival. \\
$t_{j}^{s}$ & Time that a fire arrives at node $j$, i.e. fire start time at node $j$. \\ 
$t_{j}^{m}$ & Time that the fire in node $j$ reaches its maximum size and spreads to its neighboring nodes. \\ 
$t_{j}^{e}$ & Time that the fire in node $j$ will burn out by itself. \\ 
$t_{j}^{v}$ & Time that a vehicle arrives at node $j$. \\
$l_{j}^{v}$ & The amount of time that the visiting UAV spends loitering at node $j$ after processing it.\\
$p_j$ & Collected value at node $j$. \\
\bottomrule
\end{tabular}
\label{table:notation}
\end{table}

\subsection{Exact MIP model (EM)}
\textcolor{NavyBlue}{We now present the Exact Model (EM), our MIP formulation, and describe the design and function of its constraints.}

\allowdisplaybreaks 
\small
\begin{alignat}{3}
    &(S):  \underset{}{\text{max}}\, & & \left(\sum_{j \in N_f}\ {p_j}\right) - \lambda \cdot t_h^v  \label{eq_mip_1} \\ 
   &\text{\quad \quad \ }  \text{s.t.} & & \hspace{9cm} &\notag\\
   & & & \text{\textbf{{Equations for reward collection}}} & \notag  \\
   & & & p_j \leq \tau_j - \beta_j \cdot t_j^v - \tau_j \cdot b_j & \forall \  j \in N_f \label{eq_mip_2} \\
   & & & b_j \geq y_j - M_j^3 \cdot t_j^v & \forall \  j \in N_f \label{eq_mip_3} \\
   & & & \text{\textbf{{Scheduling and routing decisions}}} & \notag  \\
   & & & \sum_{j \in N_f} {x^{k}_{h,j}} = \sum_{j \in N_f}{x^{k}_{j,h}} & \forall \ k \in K \label{eq_mip_4} \\
   & & & \sum_{j \in N_f} {x^{k}_{h,j}} \leq 1 & \forall \ k \in K \label{eq_mip_5} \\
   & & &\sum_{\substack{i\in N_f {\cup} \{h\} \\ i \neq j}}{x^k_{ij}} = \sum_{\substack{i\in N_f {\cup} \{h\} \\ i \neq j}}x^k_{ji} & \forall \ j \in N_f, \  \forall \ k \in K  \label{eq_mip_6}\\
   & & & \sum_{k \in K}\sum_{\substack{i \in N_f  {\cup} \{h\} \\ i \neq j}}{x_{i,j}^{k}} \leq 1 & \forall \ j \in N_f \label{eq_mip_7} \\
   & & & x_{i,j}^{k} = \sum_{l \in N_w}{w_{i,j,l}^{k}} & \forall i, j \in N_f\ i {\neq}j, k \in K \label{eq_mip_8} \\
   & & & 2 \cdot w_{i,j,l}^{k} \leq x_{i,l}^{k} + x_{l,j}^{k} & \forall i, j \in N_f\ i{\neq}j, k {\in} K, l{\in} N_w \label{eq_mip_9} \\
   & & &  \sum_{\substack{j \in N_f \\ j \neq i}}{x_{i,j}^{k}} =  \sum_{l \in N_w}{x_{i,l}^{k}} & \forall \ i \in N_f, \ k \in K \label{eq_mip_10} \\
   & & & \sum_{\substack{i \in N_f \\ i \neq j}}{x_{i,j}^{k}} =\sum_{l \in N_w}{x_{l,j}^{k}} & \forall \ j \in N_f, \  k \in K \label{eq_mip_11} \\
   & & & t_{h}^{v} \leq t_{max} & \label{eq_mip_12}  \\
   & & & t_h^v \geq t_j^v + d_{j,h} \cdot \sum_{k \in K}{x_{j,h}^k} - M_{j}^{13} \cdot \Bigl(1-\sum_{k \in K}{x_{j,h}^k}\Bigl)  & \forall \  j \in N_f \label{eq_mip_13} \\
   & & & t_j^v \leq l_h^v + d_{h,j} \cdot \sum_{k \in K}{x_{h,j}^k} + M_{j}^{13} \cdot \Bigl(1-\sum_{k \in K}{x_{h,j}^k}\Bigl)  & \forall \  j \in N_f \label{eq_mip_14} \\
   & & & t_j^v \geq l_h^v + d_{h,j} \cdot \sum_{k \in K}{x_{h,j}^k} - M_{j}^{13} \cdot \Bigl(1-\sum_{k \in K}{x_{h,j}^k}\Bigl)  & \forall \  j \in N_f \label{eq_mip_15} \\
   & & & t_j^v \leq t_i^v + l_i^v + \sum_{l \in N_w}\sum_{k \in K} d_{i,l}{x_{i,l}^k} + \sum_{l \in N_w}\sum_{k \in K} d_{l,j} {x_{l,j}^k} + M_{i,j}^{16} \Bigl(1-\sum_{k \in K}{x_{i,j}^k}\Bigl)  &  \forall  i,j {\in N_f}\ i{\neq}j\label{eq_mip_16} \\
   & & & t_j^v \geq t_i^v + l_i^v + \sum_{l \in N_w}\sum_{k \in K}d_{i,l}{x_{i,l}^k} + \sum_{l \in N_w}\sum_{k \in K}d_{l,j} {x_{l,j}^k} - M_{i,j}^{16} \Bigl(1-\sum_{k \in K}{x_{i,j}^k}\Bigl)  & \forall i,j {\in N_f}\ i{\neq}j  \label{eq_mip_17} \\
   & & & t_j^v \leq M_{j}^{13} \cdot \sum_{\substack{i \in N_f {\cup} \{h\} \\ i \neq j}}\sum_{k\in K}{x_{i,j}^{k}}  & \forall \  j \in N_f \label{eq_mip_18} \\
   & & & l_j^v \leq M_{j}^{13} \cdot \sum_{\substack{i \in N_f {\cup} \{h\} \\ i \neq j}}\sum_{k\in K}{x_{i,j}^{k}}  & \forall \  j \in N_f \label{eq_mip_19} \\
   & & & t_j^v - t_j^s \geq M^{20} \cdot \Bigl(\sum_{\substack{i \in N_f  {\cup} \{h\} \\ i \neq j}}\sum_{k\in K}{x_{i,j}^{k}} - 1 \Bigl)  & \forall \  j \in N_f \label{eq_mip_20} \\
   & & & t_j^v \leq t_j^e & \forall \  j \in N_f \label{eq_mip_21} \\
   & & & \text{\textbf{{Interactions between scheduling and fire progression decisions}}} & \notag  \\
   & & & M_i^{22} \cdot t_i^v \geq (1-s_i^1) & \forall \  i \in N_f \label{eq_mip_22} \\
   & & & t_i^v \leq M_i^{23} \cdot s_i^4 & \forall \  i \in N_f \label{eq_mip_23} \\ 
   & & & t_i^v - t_i^m + \varepsilon \leq M_{i}^{24} \cdot s_i^2 & \forall \  i \in N_f  \label{eq_mip_24} \\ 
   & & & t_i^m - t_i^v  \leq M^{25} \cdot (s_i^1 + s_i^3) & \forall \  i \in N_f  \label{eq_mip_25} \\ 
   & & & s_i^1 + s_i^4 =1 & \forall \  i \in N_f  \label{eq_mip_26} \\ 
   & & & s_i^4 \geq s_i^2 + s_i^3  & \forall \  i \in N_f  \label{eq_mip_27} \\ 
   & & & \sum_{j{\in}i^+}{z_{i,j}} \leq |i^+| \cdot y_i   & \forall \  i \in N_f \label{eq_mip_28} \\ 
   & & & \sum_{j{\in}i^+}{z_{i,j}} \geq |i^+| \cdot (s_i^1 + y_i -1)   & \forall \  i \in N_f \label{eq_mip_29} \\ 
   & & & \sum_{j{\in}i^+}{z_{i,j}} \geq |i^+| \cdot s_i^2  & \forall \  i \in N_f \label{eq_mip_30} \\ 
   & & & \sum_{j{\in}i^+}{z_{i,j}} \leq |i^+| \cdot (1 - s_i^3)  & \forall \  i \in N_f \label{eq_mip_31} \\ 
   & & & |j^+| \cdot y_j \geq \sum_{i \in j^+}{z_{i,j}} & \forall \  j \in N_f \label{eq_mip_32} \\
   & & & y_j \geq \sum_{k\in K}\sum_{\substack{i \in N_f {\cup} \{h\} \\ i \neq j}}{x_{i,j}^{k}}  & \forall \ j \in N_f \label{eq_mip_33} \\
   & & & \sum_{j\in N_a}{y_j} = |N_a| & \label{eq_mip_34} \\
   & & & |j^+| \cdot \sum_{i\in j^+}{q_{i,j}} \geq \sum_{i \in j^+}{z_{i,j}} & \forall \  j \in N_f \label{eq_mip_35} \\
   & & & \sum_{i\in j^+}{q_{i,j}} \leq 1 & \forall \  j \in N_f \label{eq_mip_36} \\
   & & & q_{i,j} \leq z_{i,j} & \forall \  j \in N_f, \ i \in j^+ \label{eq_mip_37} \\
   & & & t_j^s \leq t_i^m + M^{38} \cdot (1 - z_{i,j}) & \forall \  j \in N_f, \ i \in j^+ \label{eq_mip_38} \\
    & & & t_j^s \geq t_i^m - M^{38} \cdot (2 - z_{i,j} - q_{i,j} + a_j ) & \forall \  j \in N_f, \ i \in j^+ \label{eq_mip_39} \\
   & & & t_j^s \leq  M^{38} \cdot \sum_{i \in j^+}{z_{i,j}} & \forall \  j \in N_f \label{eq_mip_40} \\
   & & & \sum_{j \in N_a}{t_j^s} = 0 & \label{eq_mip_41} \\
   & & & t_j^m = t_j^s + \frac{\delta_j}{d_j} & \forall \ j \in N_f \label{eq_mip_42} \\
   & & & t_j^e = t_j^m + \frac{\delta_j}{\alpha_j} & \forall \ j \in N_f\label{eq_mip_43} \\
   & & & \text{\textbf{{Decision variable definitions}}} & \notag  \\
   & & & p_j, \ t_j^s, \  t_j^m, \  t_j^e, \  t_j^v, \  l_j^v \geq 0 & \forall \ j \in N_f\label{eq_mip_44} \\
   & & & y_{j}, \ b_{j}, \ s_{j}^c,  \in \{0,1\} & \forall \ j \in N_f, \  c \in \{1,2,3,4\}\label{eq_mip_45} \\
   & & & x_{i,j}^k, \ w_{i,j,l}^k, \ z_{i,j}, \ q_{i,j} \in \{0,1\} & \forall i, j{\in}N_f,i{\neq}j, k{\in}K, l{\in}N_w\label{eq_mip_46}
\end{alignat}
\normalsize
The model optimizes the scheduling and routing decisions for a fleet of firefighting UAVs, considering the fire spread in the region. The objective function (\ref{eq_mip_1}) maximizes the total remaining value of the region. There is also a second term in the objective function with a small enough penalty coefficient that ensures constraint (\ref{eq_mip_13}) functions properly, allowing UAVs to return to the home base immediately after completing their latest assignments. The model comprises four groups of constraints. The first group, constraints (\ref{eq_mip_2} - \ref{eq_mip_3}), defines the remaining value at nodes. The second set of equations, constraints (\ref{eq_mip_4} - \ref{eq_mip_21}), defines scheduling and routing decisions, which include assigning vehicles to fire nodes, determining the order in which fire nodes are visited by vehicles, selecting water resources between the visited nodes, and setting vehicle arrival times and loitering times at nodes. The third set of equations, constraints (\ref{eq_mip_22} - \ref{eq_mip_43}), links the fire spread or progression (such as degradation, amelioration, and spreading to the neighboring areas) with the scheduling and routing decisions. The fourth and final set of equations, constraints (\ref{eq_mip_44} - \ref{eq_mip_46}), define the decision variables. 

Note that the model incorporates nine big-M parameters that support the functionality of their respective constraints. We use the notation $M_j^k$, where $k$ corresponds to the constraint ID where the big-M parameter is initially introduced in the model, and $j$ represents the node $j$. This is because we use different big-M values for each node $j$ in certain cases. We provide the necessary theoretical background to comprehend the roles of these big-M parameters and specify their minimum values in Section B of the Supplementary Material. Next, we delve into the details of the main constraint groups of our model.

\subsubsection{Maximizing the collected remaining value}
Constraint (\ref{eq_mip_2}) determines the collected value, $p_j$, for each node $j \in N_f$ based on three cases. In case (1), no fire spreads to node $j$, the model sets $t_j^v$ and $b_j$ to 0, and the initial value of node $j$, $\tau_j$, is collected in full. In cases (2) and (3) a fire spreads to node $j$ and its initial value degrades over time until the fire is extinguished. In case (2), a vehicle processes the fire at node $j$, i.e. $t_j^v>0$ and $b_j=0$, and the remaining value at $t_j$ is collected. In case (3), the fire is not processed, i.e. $t_j^v=0$ and $b_j=1$, resulting in zero remaining value as the fire burns everything until it burns out on its own. The MIP model prioritizes avoiding fire spread to the nodes, i.e. setting $y_j=0$ for each node $j$, as case (1) is the ideal scenario in terms of maximizing the remaining value with the objective function (\ref{eq_mip_1}). If a fire spreads to node $j$, then the model prioritizes assigning a vehicle to process it as case (2) produces a higher value collection than case (3). We next demonstrate the properties of the model that enable it to handle this prioritization. We provide the propositions here, with their proofs available in Section C of the Supplementary Material.

\begin{proposition}
 $p_j$ is maximized when $y_j=0$. That is, the collected value at node $j$ is maximized when no fire spreads to it (case 1).
 \label{proposition:1}
\end{proposition}

Having no fire ($y_j=0$) leads to the highest collected value at node $j$, as per Proposition \ref{proposition:1}. The value of $y_j$ can be set to 0 only if $z_{i,j}=0$ for all $i \in j^+$ by constraint (\ref{eq_mip_32}). The model sets $z_{i,j}=0$ only if adjacent nodes are fire free or their fires are processed promptly before they spread to $j$. Therefore, the model will always attempt to process fires in $i \in j^+$ to prevent $z_{i,j}$ from being non-zero and to maximize $p_j$. Yet, it may not always be feasible to prevent fire spread due to limited server availability. If a fire spreads from an adjacent node $i$ to node $j$, constraints (\ref{eq_mip_22} - \ref{eq_mip_31}) set $z_{i,j}=1$, mandating $y_j=1$ in constraint (\ref{eq_mip_32}). Subsequently, the model determines whether to assign a vehicle to node $j$  for fire processing by setting $t_j^v>0$, or not. 

\begin{proposition}
$p_j$ is maximized if $t_j^v>0$ when $y_j=1$.
 \label{proposition:2}
\end{proposition}

When there is a fire at node $j$ ($y_j=1$), assigning a vehicle ($t_j^v>0$) maximizes the collected value at that node.

\subsubsection{Scheduling and routing decisions}
Constraints (\ref{eq_mip_4} - \ref{eq_mip_11}) determine vehicle-job assignments and provide flow balance, while  constraints(\ref{eq_mip_12} - \ref{eq_mip_21}) set the arrival times at nodes. 

Constraint (\ref{eq_mip_4}) ensures that all vehicles leaving the base must return to it, while constraint (\ref{eq_mip_5}) allows each vehicle to leave the base only once. Constraint (\ref{eq_mip_6}) ensures that a vehicle that visits a node must also leave the node. Constraint (\ref{eq_mip_7}) states that a region can only be visited once or not at all. Constraint (\ref{eq_mip_8}) requires a vehicle that moves from node $i$ to $j$ to select a water resource for refilling its tank, otherwise, no water resource should be visited. Constraint (\ref{eq_mip_9}) ensures that once a water resource is selected between nodes $i$ and $j$, it must be visited as an intermediate point during the travel from node $i$ to $j$. Constraints (\ref{eq_mip_10} - \ref{eq_mip_11}) complete constraints (\ref{eq_mip_8} - \ref{eq_mip_9}) by ensuring that when a vehicle travels from node $i$ to $j$, it must first fly from node $i$ to a water resource and then from the water resource to node $j$. 

Constraints (\ref{eq_mip_12}-\ref{eq_mip_13}) establish a time limit for the operation using the arrival time of the last vehicle returning to the base. Constraints (\ref{eq_mip_14} - \ref{eq_mip_17}) determine the arrival times of the nodes. In our application, we omit the service time. However, if desired, service durations can either be included in travel durations or incorporated as distinct parameters within the model formulation.

The UAVs start at the base with full tanks. When flying between nodes, excluding the base, travel time includes refilling at a water resource. The flow variables are summed over all vehicles and water resources, but the flow balance constraints ensure only one vehicle to visit a node and only one water resource is used for refilling. UAVs can loiter between tasks to benefit the overall objective. For example, if a UAV finishes a task, it might loiter even if there is an existing fire if it knows a fire will soon spread to a more valuable location. Constraint (\ref{eq_mip_18}) sets arrival times at unvisited nodes to 0. Constraint (\ref{eq_mip_19}) prohibits loitering at unvisited nodes. Constraint (\ref{eq_mip_20}) ensures that a node can only be visited after the fire has started. Constraint (\ref{eq_mip_21}) ensures arrival before the fire burns out.

The model differs from the formulation of the basic VRP in two ways: (1) each vehicle must visit a water resource to refill its tank when moving from one job to another, and (2) the arrival times at nodes need to be specified as they are linked to the fire spread decisions made by constraints (\ref{eq_mip_22} - \ref{eq_mip_43}). Explicitly determining arrival times eliminates the need for sub-tour elimination constraints.  

\subsubsection{Interactions between scheduling - routing decisions and fire arrivals}
\label{section:interactions}
Constraints (\ref{eq_mip_22}) to (\ref{eq_mip_43}) capture the inter-dependencies between scheduling decisions and fire progression. Constraints (\ref{eq_mip_22}) to (\ref{eq_mip_31}) determine if a fire spreads to adjacent nodes. Constraint (\ref{eq_mip_32}) ensures that fire spread to node $j$ triggers a fire arrival at node $j$. Constraint (\ref{eq_mip_33}) specifies that a node can only be visited if it has a fire. Constraint (\ref{eq_mip_34}) introduces the initial active fire conditions. Constraints (\ref{eq_mip_35}) to (\ref{eq_mip_40}) determine the start time of new fires based on spread from adjacent nodes. Constraint (\ref{eq_mip_41}) sets the start times of the initially active fires to zero. Constraint (\ref{eq_mip_42}) computes the time for a fire to reach its maximum size (covering the entire grid and spreading to adjacent nodes), while constraint (\ref{eq_mip_43}) computes the time for a fire to burn out on its own. 

\begin{figure}[ht] 
\centering
\includegraphics[scale=0.7]{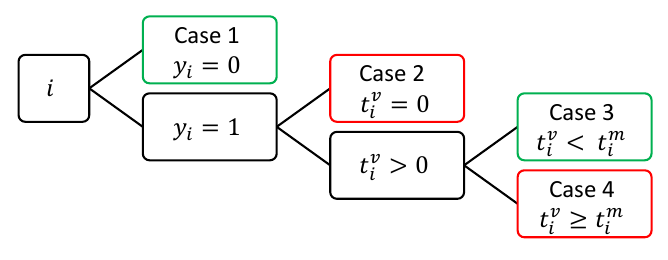}
\caption{Cases triggering fire spread in node $i$}
\label{figure:fire_soread_cases}
\end{figure}

Next, we explore the properties that enable constraints (\ref{eq_mip_22} - \ref{eq_mip_43}) function properly. Constraints (\ref{eq_mip_22}) - (\ref{eq_mip_31}) define four cases to capture different scenarios based on the fire conditions of the nodes and scheduling and routing decisions, illustrated in Figure \ref{figure:fire_soread_cases}, where green and red borders indicate no fire spread and fire spread, respectively.
\textbf{Case 1:} No fire in node $i$ ($y_i=0$) and no fire spreads to adjacent nodes.
\textbf{Case 2:} Fire in node $i$ ($y_i=1$), no vehicle processes it ($t_i^v=0$), and the fire spreads to adjacent nodes.
\textbf{Case 3: }Fire in node $i$ ($y_i=1$), a vehicle processes it ($t_i^v>0$) before  it spreads ($t_i^v<t_i^m$), preventing fire spread to adjacent nodes.
\textbf{Case 4:} Fire in node $i$ ($y_i=1$), a vehicle processes it ($t_i^v>0$) after it spreads ($t_i^v\geq t_i^m$), and the fire spreads to adjacent nodes.

In case (1), the model ensures no fire spreads from node $i$ to $j \in i^+$ per Proposition \ref{proposition:3}. 

\begin{proposition}
$z_{i,j}$ is set to $0$ for all $j \in i^+$ when $y_i=0$.
\label{proposition:3}
\end{proposition}

In the event of a fire ($y_i=1$), whether cases 2, 3, and 4 occur depends on the assignment of vehicles to the fire. If no vehicle is assigned, it results in $s_i^1=1$ and $s_i^4=0$, whereas assigning a vehicle results in $s_i^4=1$ and $s_i^1=0$. In essence, when $s_i^4=1$, it initiates the possibilities of Cases 3 and 4, which are then determined by $s_i^2$ and $s_i^3$. These connections are elaborated upon in the subsequent proofs.

In case (2), the model ensures that the fire spreads from node $i$ to all $j \in i^+$, as per Proposition \ref{proposition:4}. 

\begin{proposition}
When $y_i=1$ and $t_i^v=0$, $z_{i,j}$ is set to 1 for all $j \in i^+$.
\label{proposition:4}
\end{proposition}

In case (3), the model prevents fire sprad from node $i$ to $j \in i^+$, as per Proposition \ref{proposition:5}. 

\begin{proposition}
When $y_i=1$ and $0<t_i^v<t_i^m$, $z_{i,j}$ is set to 0 for all $j \in i^+$.
\label{proposition:5}
\end{proposition}

In case (4), the model ensures that the fire spreads from node $i$ to all $j \in i^+$, as per Proposition \ref{proposition:6}. 

\begin{proposition}
When $y_i=1$, and $t_i^v \geq t_i^m > 0$, $z_{i,j}$ is set to 1 for all $j \in i^+$. 
\label{proposition:6}
\end{proposition}

Once a fire spread decision is made, the next task is to determine the time of the spread. For a node $j$, constraints (\ref{eq_mip_22} - \ref{eq_mip_31}) separately evaluate the spread decisions for each node $i$ adjacent to node $j$. Therefore, fire can spread from multiple adjacent nodes to the same node $j$. This does not harm the model results as long as the start time of the arriving fire is set to the earliest spread time. The earliest spread time is specified by Equation (\ref{equation:min_spread_time}). 

\begin{equation}
    t_j^s\ =
    \left\{
        \begin{array}{ll}
            0 & \text{if} \ a_j=1 \  \text{or} \  \sum\limits_{i \in j^+}{z_{i,j}} = 0 ,\\
             \min\limits_{i \in j^+}{\Bigl\{t_i^m:z_{i,j} >0 \Bigl\} } & \text{otherwise.} 
    \end{array}
    \right.
\label{equation:min_spread_time}
\end{equation}

If there is an active fire in node $j$ initially ($a_j=1$), or no fire spreads to region $j$ from its adjacent nodes ($\sum\limits_{i \in j^+}{z_{i,j}} = 0$), then no fire arrives at node $j$ and $t_j^s$ is set to zero. If node $j$ does not have an initial fire but a new fire spreads from its adjacent nodes, the minimum $t_i^m$ among the adjacent nodes that have non-zero $z_{i,j}$ values is assigned to $t_j^s$. In other words, it sets the earliest time fire starts at node $j$, triggered by the first neighboring node reaching the critical fire size. This expression is nonlinear and requires linearization for the MIP model, as established by proposition \ref{proposition:7}.

\begin{proposition}
Constraints (\ref{eq_mip_35} - \ref{eq_mip_40}) linearize Equation (\ref{equation:min_spread_time}).
\label{proposition:7}
\end{proposition}

{\color{NavyBlue}
\subsubsection{Model Complexity}
We analyze the model size using big-\(O\) notation in Section D of the Supplementary Material. Since $|K|$, $|N_w|$, and $\Delta$ (the maximum number of neighbors per node) are typically fixed and small in real-world applications, the overall model complexity simplifies to $O(|N_f|^2)$. See the Supplementary Material for a detailed breakdown of variables and constraints.
}

{\color{NavyBlue}
\subsection{Hybrid Model: Exact Model-Guided Refill Lazy Model (E-RLM)}
Solving the EM with Gurobi poses  performance challenges (see Sections \ref{comp_experiments} and \ref{case_study}). The solver quickly identifies high-quality feasible solutions but spends almost the entire time limit attempting to improve the upper bound and prove optimality, due to weak LP relaxations and limited branch pruning. We address the performance problem with a hybrid model. We first construct a smaller MIP model by relaxing the refill-related constraints of EM, referred to as the Refill Lazy EM (RLM). Building on this, we develop the EM-Guided Refill Lazy Model (E-RLM), a hybrid model that leverages EM’s solution and RLM’s relaxed formulation within a branch-and-cut framework incorporating dynamic constraint generation. 

\subsubsection{Refill-Lazy Exact Model (RLM)}
\label{section_rlm}

EM enforces water resource visits between consecutive tasks through Constraints (8) and (9), and accounts for the associated travel times in Constraints (16)–(17). These constraints introduce complexity in managing temporal coupling. By relaxing them, we derive RLM. Relaxing time-window-related constraints is a recognized and beneficial strategy in the literature (e.g., \citet{vidal2015time}; \citet{LAU2003559}).

Specifically, to obtain RLM, we omit Constraints (\ref{eq_mip_8}) and (\ref{eq_mip_9}), removing the requirement to visit a water resource between consecutive tasks. We also omit Constraint (\ref{eq_mip_16}), eliminating the upper bound on arrival times, and relax Constraint (\ref{eq_mip_17}) as Constraint (\ref{equation:relaxed_lower_arrival_}), which now sets arrival times based on the preceding node's visit time and travel duration, excluding the travel time to and from a water resource. The remaining constraints and objective function are identical to those in EM.

\begin{equation}
   t_j^v \geq t_i^v + l_i^v + \sum_{k \in K}d_{i,j}{x_{i,j}^k} - M_{i,j}^{16} \Bigl(1-\sum_{k \in K}{x_{i,j}^k}\Bigl)  \quad \quad  \forall i,j {\in N_f}\ i{\neq}j
\label{equation:relaxed_lower_arrival_}
\end{equation}

Our branch-and-cut with dynamic constraint generation proceeds as follows. We solve RLM and invoke a lazy constraint callback to dynamically enforce the omitted water resource constraints (\ref{eq_mip_8}, \ref{eq_mip_9}) and time coupling constraints (\ref{eq_mip_16}, \ref{eq_mip_17}) whenever they are violated by an incumbent solution. When violations are detected, the corresponding constraints are added as lazy cuts, and the solver resumes branching. This process repeats until a solution satisfying all EM constraints is found. By delaying the introduction of these constraints, we leverage the reduced model size in the early stages of the run. This strategy is well-established in the literature; for example, in the TSP, subtour elimination constraints are added only when violated.

A critical challenge arises with this approach. In branch-and-cut with lazy constraints, the solver progresses by making decisions based on RLM. As a result, branches that appear suboptimal under RLM may be pruned, even though they could lead to the optimal solution under EM. For example, a promising branch (A) may be explored, while another (B) is pruned early. If refilling constraints added later render branch A infeasible, the solver cannot revisit branch B even if it contains the true EM-optimal solution. We address this limitation through a hybrid model, which we introduce next.

\subsubsection{Algorithmic Steps of E-RLM}
We develop E-RLM  as a two-stage hybrid algorithm that combines the high-quality solution generation of EM with the strong bound improvement capabilities of RLM with dynamic constraint generation. Let $T$ denote the total time limit for E-RLM, and $T^{\text{EM}} = rT$ the time allocated to EM, with $r \in [0,1]$. Let $T^{\text{now}}$ represent the current elapsed runtime, $\bar{\epsilon}$ the optimality gap threshold, $x^*$ the final solution, and $\epsilon^*$ its corresponding optimality gap. The best solutions and corresponding gaps obtained from EM and RLM are denoted by $(x^{\text{EM}}, \epsilon^{\text{EM}})$ and $(x^{\text{RLM}}, \epsilon^{\text{RLM}})$, respectively. The steps of E-RLM are as follows:

\begin{spacing}{1.2}
\makeatletter
\renewcommand{\fnum@algorithm}{} 
\makeatother
\begin{algorithm}[ht]
\caption*{\textbf{Exact Guided Refill Lazy Model (E-RLM)}}  
\label{alg:e-rlm}
\begin{algorithmic}[]
\small
\State \textbf{Step 1: Initialize}
\State Set $\bar{\epsilon}$, $T$, $r$ and compute $T^{\text{EM}}$.
\State Initialize $x^* \leftarrow \emptyset$, $\epsilon^* \leftarrow \emptyset$, and $T^{\text{now}} \leftarrow 0$.
\State Proceed to Step 2.
\State \textbf{Step 2: Solve EM}
\State Run EM until $\epsilon^{\text{EM}} \leq \bar{\epsilon}$ or $T^{\text{now}} > T^{\text{EM}}$.
\If{$\epsilon^{\text{EM}} \leq \bar{\epsilon}$ and $T^{\text{now}} \leq T^{\text{EM}}$}
    \State Set $x^* \leftarrow x^{\text{EM}}$, and $\epsilon^* \leftarrow \epsilon^{\text{EM}}$.
    \State Proceed to Step 4.
\Else
    \State Proceed to Step 3.
\EndIf
\State \textbf{Step 3: Solve RLM}
\State Run RLM with $x^{\text{EM}}$ as warm start using lazy branch-and-cut, until $\epsilon^{\text{RLM}} \leq \bar{\epsilon}$ or $T^{\text{now}} > T$.
\State Set $x^* \leftarrow x^{\text{RLM}}$, and $\epsilon^* \leftarrow \epsilon^{\text{RLM}}$
\State Proceed to Step 4.
\State \textbf{Step 4: Termination}
\State \textbf{Output} $x^*$, $\epsilon^*$, and $T^{\text{now}}$.
\State Terminate E-RLM.
\normalsize
\end{algorithmic}
\end{algorithm}
\end{spacing}

As outlined in the steps of E-RLM, the model begins by initializing key parameters and variables (Step 1). It then first runs EM for a limited time (Step 2). If the optimality gap meets the threshold, the model terminates early. Otherwise, it proceeds to RLM (Step 3), which uses the EM solution as a warm start and applies a branch-and-cut method with lazy constraint enforcement. RLM runs until the optimality gap meets the threshold or the overall time limit is reached.
}

\section{Computational experiments}
\label{comp_experiments}
We develop a Python program utilizing the GUROBI Optimizer 10.0.1 \citep{gurobi2024} through the gurobipy API to solve our mathematical model. We design this program as an application that can be executed by defining inputs on a spreadsheet file, eliminating the need to interact with the underlying code-base. Users can employ the application to generate new problem instances and/or search for optimal solutions for their problem instances. The code-base, along with our problem instances in this paper, is openly shared as a GitHub repository\footnote{\href{https://github.com/edasdemirlab/job-scheduling-fire-2024.git}{https://github.com/edasdemirlab/job-scheduling-fire-2024.git}: The repo is private to keep the code confidential before publication. We have included our codebase as a supplementary file to facilitate its review by the referees.}. We utilize this program to conduct our computational experiments. The computer we use is a 2-core 28-thread Dell work station running Linux Centos 7 with a processor Intel Xeon Gold 6330 CPU, @28 x2.10GHz and 1 TB usable RAM \citep{CCR2024}. 

\subsection{Problem instance}
\begin{figure}[h] 
\centering
\includegraphics[scale=1]{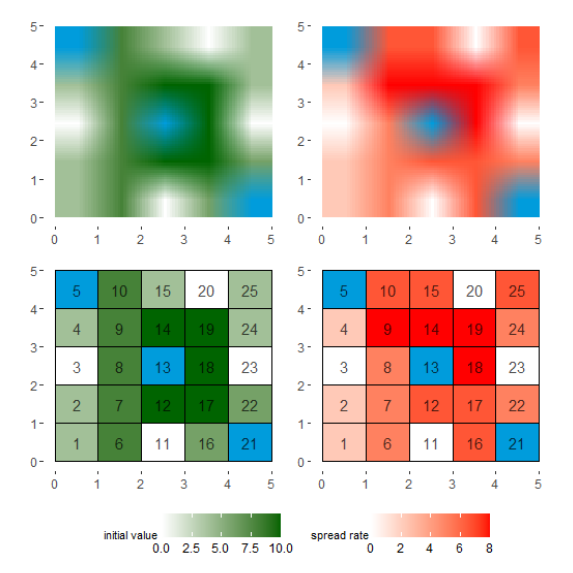}
\caption{5x5 problem instance}
\label{figure:5x5_problem-instance}
\end{figure}

We generate an empirical problem for the experimental study in this section. We have additional practical problem instances in the case study section (see Section \ref{subssec:realistic_scenarios}). Figure \ref{figure:5x5_problem-instance} illustrates the initial value and spread maps of the region, along with their grid representations.

The spread rates, expressed in km/h, range from 0 to 8, where 0 denotes blocking nodes with no fire spread and 8 indicates the maximum spread rate. The fire amelioration rate spans from 0 to 4 km/h. The initial values of the grids range from 0 to 10. The value degradation rate of a grid is calculated by dividing its initial value by time it takes for a fire to burn down naturally. This encompasses reaching the maximum size and returning to a size of 0. For a justification of all the numerical values here, please refer to Section \ref{subssec:realistic_scenarios}. For the specific instance here, the rates within the above ranges are arbitrarily assigned.

\subsection{Experimental settings}
\begin{figure}[h] 
\centering
\includegraphics[scale=0.7]{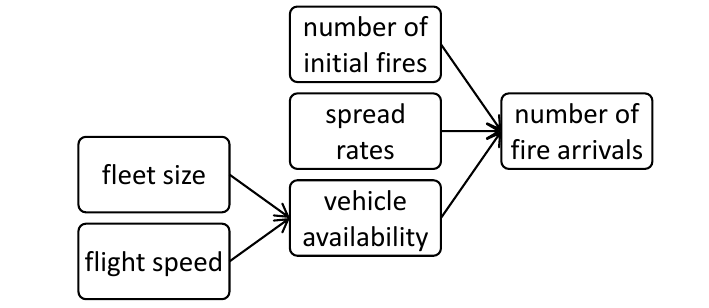}
\caption{Parameters impacting number of fire arrivals}
\label{figure:computation_experiments_factors}
\end{figure}

We observe that the number of fire arrivals plays a pivotal role in both the objective function and solution time. We report computational results for this insights in the next subsection. Although the number of fire arrivals is not a problem parameter that can be directly controlled, its value is set by four controllable parameters: (1) fleet size and (2) flight speed, which together determine vehicle availability, and (3) the number of initial fires and (4) the spread rates of grids. We illustrate this in Figure \ref{figure:computation_experiments_factors}. 

The parameters (1) and (2) are straightforward to adjust. We consider two different sets of vehicles, K=\{1, 2\} and K=\{1,2,3,4\}, and two different vehicle speeds, 60km/h and 120km/h. However, adjusting parameters (3) and (4) require more careful consideration. A higher number of initial fires typically results in more jobs to process and an increased potential for further fire spread, which, in turn, raises the number of total job arrivals. Moreover, the specific nodes where these initial fires are located become critical, as this is where parameter (4), regional fire spread rates, starts to exert its influence. For instance, if the initial fires occur in nodes with higher spread rates, the likelihood of more job arrivals is greater.

To ensure that our investigation into the impact of parameters (3) and (4) remains unbiased and comprehensive, we take a systematic approach. Our problem instance has 17 potential nodes that can have fire initially (see Figure \ref{figure:5x5_problem-instance}). We explore various combinations of these initial fire locations. Specifically, we consider 20 different combinations for each number of initial fires. For example, if we start with 2 initial fires, this results in 136 possible combinations to allocate the initial fires to 17 candidate nodes. Given the extensive nature of this experimentation, we randomly sample 20 allocation options for each initial number of fires. In cases with fewer than 20 possibilities, such as when the initial number of fires is 1 (which leads to 17 options), we include all available options.

For each vehicle size and speed pair, the sampled combinations lead to 315 runs, totaling 1260 distinct executions of the model. \textcolor{NavyBlue}{For termination criteria, the optimality gap threshold $\bar{\epsilon}$ is set to 0.03 for both EM and E-RLM. The time limit for EM is fixed at 3,600 seconds for all instances. For E-RLM, the total time limit $T$ is instance-specific and determined based on the number of initial fires, given by $T = 0.2 \times 60 \times |{N_a}|$, where $|{N_a}|$ denotes the number of initial fires. This corresponds to allocating 12 seconds per active fire at start, with a maximum of 204 seconds when $|{N_a}| = 17$. In the E-RLM setting, 60\% of $T$ is reserved for solving EM (i.e., $r = 0.6$)}. These arbitrary choices are sufficient in all experiments, as shown in the next section.

\subsection{Results}
\begin{figure}[ht] 
\centering
\includegraphics[scale=0.7]{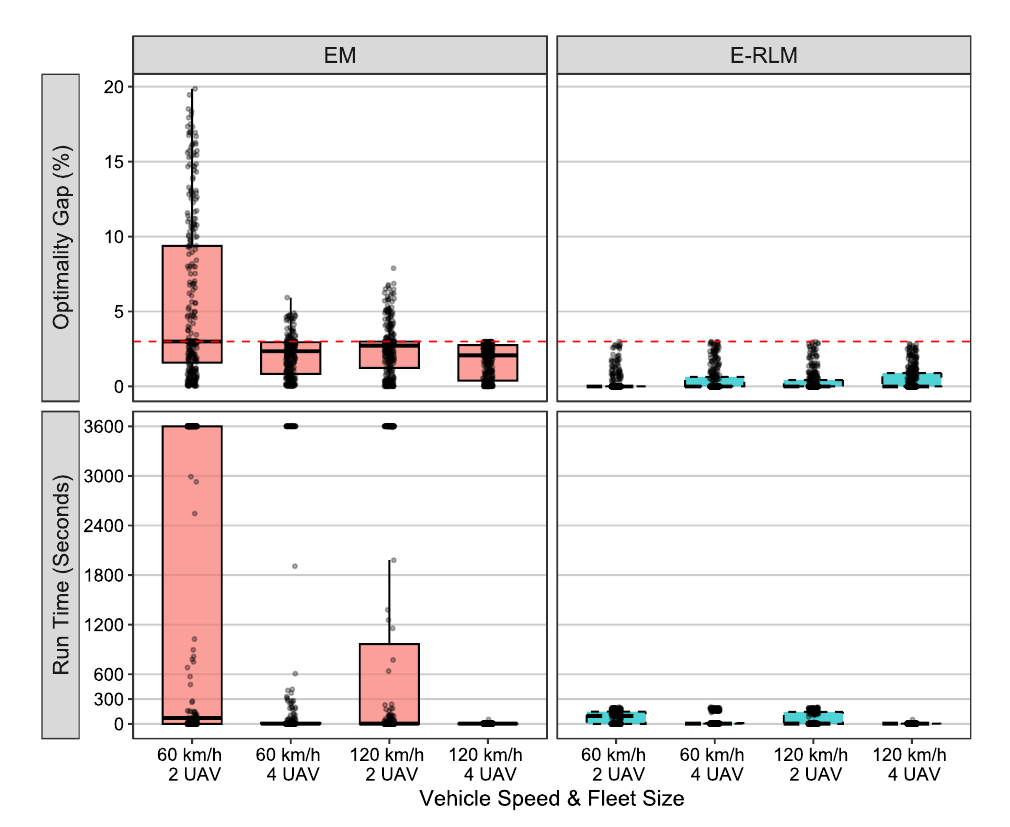}
\caption{\textcolor{NavyBlue}{Comparison of EM and E-RLM in terms of optimality gap and run time across fleet size,  speed setting, and initial fire conditions.}}
\label{figure:comparison_em_erlm}
\end{figure}

\textcolor{NavyBlue}{
In Figure \ref{figure:comparison_em_erlm}, we present a 2×2 plot to compare EM and E-RLM in terms of optimality gap and runtime, with rows for performance metrics and columns for solution models. Each subplot contains four boxplots, one for each fleet size and speed configuration, each summarizing 315 model runs generated using distinct initial fire scenario.}

\textcolor{NavyBlue}{Performance outcomes consistently favor E-RLM. EM performs poorly under constrained fleet conditions, exhibiting large gaps and long runtimes—except in the 120 km/h, 4-UAV setting, where it solves all 315 instances optimally. In contrast, E-RLM consistently satisfies the 3\% optimality gap criterion across all 1,260 runs, regardless of fleet configuration and initial fire condition. It solves every instance with an average runtime of 47 seconds and a standard deviation of 70.}

\begin{figure}[ht] 
\centering
\includegraphics[scale=1]{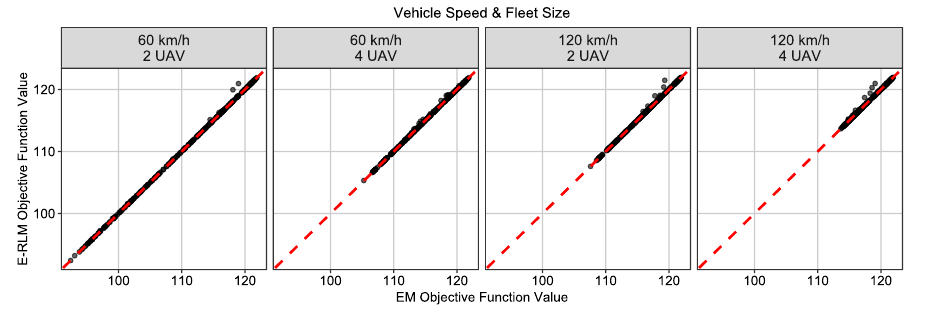}
\caption{\textcolor{NavyBlue}{Instance-wise comparison of EM and E-RLM objective values across four parameter settings}}
\label{figure:obj_em_erlm}
\end{figure}

\textcolor{NavyBlue}{We then examine the objective function values produced by EM and E-RLM. Figure \ref{figure:obj_em_erlm} presents a pointwise comparison across four parameter settings, each with 315 runs. The x-axis shows EM results, while the y-axis shows E-RLM results. The dashed red line indicates the 1-to-1 reference—instances where both models yield the same objective value lie on this line. E-RLM matches or outperforms EM in all 1,260 instances, demonstrating that the warm-start strategy we use in E-RLM effectively mitigates the risk of prematurely pruning branches containing high-quality solutions (see Section \ref{section_rlm} for our discussion of this issue). Also, the near overlap between EM and E-RLM values supports our motivation for the hybrid model E-RLM: EM often finds optimal solutions but fails to prove optimality, a gap E-RLM addresses. As vehicle speed and fleet size increase, the data points shift toward the upper right, indicating greater value retention and reduced variability across fire scenarios. This is consistent with expectations under enhanced suppression capacity.}

\begin{figure}[ht] 
\centering
\includegraphics[scale=1]{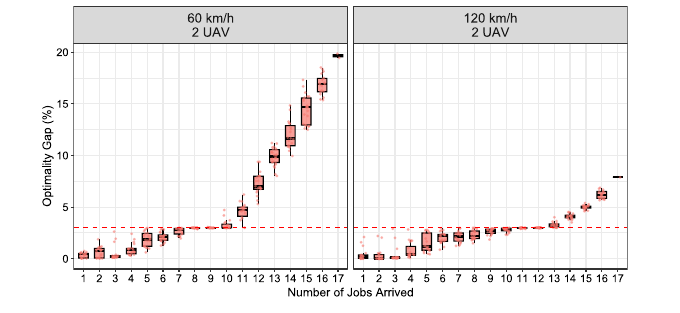}
\caption{\textcolor{NavyBlue}{Optimality gaps of EM as a function of the number of job arrivals for two fleet configurations: 2 UAVs at 60 km/h and 2 UAVs at 120 km/h}}
\label{figure:jobs_gap_em}
\end{figure}

\textcolor{NavyBlue}{Finally, we examine the relationship between the number of job arrivals and EM’s computational performance. The results are presented in Figure \ref{figure:jobs_gap_em}, which includes two columns for different fleet configurations: 2 UAVs at 60 km/h and 2 UAVs at 120 km/h. Each configuration includes 315 runs with job (fire) arrivals ranging from 1 to 17, with multiple instances per value. For each job count on the x-axis, a box plot on the y-axis displays the corresponding optimality gaps. The red dashed horizontal line marks the 3\% termination criterion. The results show that as the number of job arrivals increases, EM’s performance deteriorates, particularly under limited fleet capacity, reflected by rising optimality gaps. In the 60 km/h, 2-UAV case, run times often reach the 3,600-second limit when job arrivals exceed approximately 11, which corresponds to about 65\% of all fire-prone nodes.}

\textcolor{NavyBlue}{In summary, the results confirm the computational advantage of E-RLM over EM across diverse scenarios. E-RLM consistently satisfies the optimality gap threshold with significantly lower runtimes and greater robustness to fleet capacity and initial fire conditions. In the next section, we explore more realistic scenarios to gain deeper insights into model performance.}

\section{Case study}
\label{case_study}
\subsection{Generating realistic scenarios}
\label{subssec:realistic_scenarios}

California has experienced some of the most severe wildfire damage in the U.S. in recent years.  Climate change exacerbates wildfire seasons worldwide, causing significant losses in heavily urbanized areas like California. These losses are concentrated in the wildland-urban interface (WUI), where wildfires affect property and lives \citep{radeloff2018rapid}. The SILVIS lab at the University of Wisconsin-Madison is the well-accepted authority on categorizing land regions in the US into 13 WUI classes based on housing and vegetation density, and proximity to dense vegetation \citep{silvislab}. Housing density is defined as housing units per sq. km, and vegetation density is the percentage of land covered by vegetation.  an area is considered to be `close to veg dense area' if it is within 2.4 km of an area with a vegetation density \textgreater{}= 75\%.  The Table \ref{tab:wui-categories}.  defines the 13 land classes plus water.

\begin{table}[htp]
\centering
\begin{tabular}{|l|c|c|c|}
\hline
\textbf{SILVIS category} & \textbf{Housing density} & \textbf{Vegetation density} & \textbf{Close to veg dense area} \\ \hline
very\_low\_dens\_veg & \textless{}6 & \textgreater{}=50 & no \\ \hline
very\_low\_dens\_no\_veg & \textless{}6 & \textless{}=50 & no \\ \hline
high\_dens\_intermix & \textgreater{}= 741 & \textgreater{}50 & no \\ \hline
high\_dens\_interface & \textgreater{}=741 & \textless{}=50 & yes \\ \hline
high\_dens\_no\_veg & \textgreater{}=741 & \textless{}=50 & no \\ \hline
uninhabited\_no\_veg & 0 & \textless{}=50 & no \\ \hline
uninhabited\_veg & 0 & \textgreater{}50 & no \\ \hline
med\_dens\_interface & 50-741 & \textless{}=50 & yes \\ \hline
med\_dens\_intermix & 50-741 & \textgreater{}50 & no \\ \hline
med\_dens\_no\_veg & 50-741 & \textless{}=50 & no \\ \hline
low\_dens\_interface & 6-50 & \textless{}=50 & yes \\ \hline
low\_dens\_no\_veg & 6-50 & \textless{}=50 & no \\ \hline
low\_dens\_intermix & 6-50 & \textgreater{}50 & no \\ \hline
water & N/A & N/A & N/A \\ \hline
\end{tabular}
\caption{WUI categories as defined by the SILVIS lab \citep{silvislab}}
\label{tab:wui-categories}
\end{table}

In California, census tracts or districts are typically dominated by one of the SILVIS categories. This means that some areas are primarily composed of water, while others are dominated by the \textit{high\_dens\_intermix} SILVIS type, and so forth. To simulate realistic scenarios, we create square grid-based cases representing the landscape types. We achieve this by selecting a predefined category as the default type for the grid cells, ensuring most of the grid is covered by that land type. Then a few areas are randomly assigned other categories, including water and fire-proof. Fire-proof nodes belong to set $N_p$ in our model and act as barriers, preventing fire spread, similar to roads or human-made fire blocks. By generating diverse cases with different default categories, we can evaluate how our model handles various landscape compositions.

To determine the size of the entire region and the area each node covers, we start by considering the area a single UAV can extinguish in one pass. \textcolor{NavyBlue}{Manned air tankers, with capacities ranging from 3,000 to 35,000 liters, are well-established in firefighting. While UAVs like the K-MAX currently carry up to 2,000 liters, we assume a 10,000-liter capacity in our model, considering advancements in UAV technology and current manned aircraft capabilities.} Analyzing wildfires like Veyo West, Red Cliffs National Conservation, and Cottonwood Trail, it was found that approximately 39 gallons of water are needed to extinguish one acre of land, equivalent to 36,322 liters per square kilometer \citep{water_for_wf}. Assuming a UAV carries 10,000 liters of water, it can effectively extinguish about 0.28 square kilometers in a single pass. Simplifying, we consider one UAV capable of extinguishing 0.25 square kilometers. Therefore, a fleet of 4 UAVs can cover 1 square kilometer. Based on this, we \textcolor{NavyBlue}{uniformly} designate each node in our grid as 1 square kilometer, representing the \textcolor{NavyBlue}{maximum} area a team of extinguishable in one pass by a team of 4 UAVs. \textcolor{NavyBlue}{Since the UAV team cannot partially release water—once the tank is opened, all the water is discharged—they release their full water load regardless of the fire suppression need (fire size) within the node.}

Our model uses parameters defined in Table \ref{table:notation}. The initial value of a node, $\tau$, is set to 0 for water nodes and -1 for fire-prone nodes. For nodes with no housing density, $\tau$ is set to 1, acknowledging the land's value even without urbanization. For housing density less than $<6$, $\tau$ is randomly assigned between 1 and 2. For density between 6 and 50, $\tau$ is randomly assigned between 2 and 4. For housing density between 50 to 741, $\tau$ is randomly assigned between 4 and 10. For housing density of $741$ or more, $\tau$ is set to 10. Thus, land value increases with housing density.

Next, we consider the fire degradation rate $d$, which is analogous to the fire spread rate. Acknowledging that fire spred faster in densely vegetated areas, we establish a range of 0.4 to 8 km/h based on typical fire spread rates documented in \citet{cruz201910}. While eucalypt forests can spread up to 10.5 km/h, other wild lands have spread rates around 6 km/h. To account for various wild lands, we choose a moderate upper bound of 8 km/h. Similarly, since the lower end of the fire spread rate ranges from 0.2 to 0.6 km/h, we adopt 0.4 km/h as the lower bound.


The value of $d$ in each node is randomly set between 0.4 and 4.2 if the vegetation density is $\leq 50\%$, and between 4.2 and 8 if the vegetation density is $>50\%$. Assuming that a fire that spreads through a node fast also dies down fast, the fire amelioration rate, $\alpha$, is set to be a random value between $d-0.5$ and $d+0.5$, also in km/h. Introducing this element of randomness accommodates various influencing factors such as weather conditions, wind speed, the flammability of urban structures, and vegetation type, which collectively impact both $d$ and $\alpha$.

Now, given that $d$ is in units of km/h, it takes $\frac{1}{d}$ hours for the fire to reach an area of 1 km, or in other words, the entire area of the node. It then takes $\frac{1}{\alpha}$ hours for the fire to naturally extinguish, assuming no intervention from UAVs. Then, the total time the fire takes to burn through a node is the sum of these two periods, given by $\frac{1}{d} + \frac{1}{\alpha}$. Assuming that the initial value of the node, $\tau$, is zero by the time the fire burns through the node, the value degradation rate, $\beta$, is calculated as $\tau$ divided by the total fire time, simplified as $\frac{\tau \alpha d}{\alpha + d}$.

\subsection{Scenarios}
\label{subsec:scenarios}

We generate seven different scenarios for our case study, each on a 7x7 grid covering just under 50 square kilometers. Our goal is to examine the model's performance across diverse landscapes and WUI types in California. Each scenario's default node type is based on its SILVIS category, with several randomly located nodes assigned to other SILVIS types. For example, a high-density intermix scenario will mainly consist of high-density intermix nodes but will also include nodes from other 

Each scenario includes 1 UAV base at node 1, 2-4 water resource nodes, and nodes with initial fires, with a 20\% chance for each node to be an ignition point. Six scenarios are primarily land-based with different characteristics, while the seventh is predominantly water with some land masses. The land-based scenarios feature various SILVIS categories as detailed below:

\begin{itemize}
    \item \textbf{Scenario 1 (High-density intermix)}: High population and dense vegetation. An example is parts of Alameda County, which includes cities like Oakland and Berkeley. They have a high population density and a mix of urban and suburban areas with a significant amount of vegetation. An aerial view of Alameda County is presented in Figure E.1 of the Supplementary Material. Figure \ref{fig:scenario_1_problem_defnition} in Section \ref{sec:demo_scenario} illustrates the associated generated scenario.

    \item \textbf{Scenario 2 (High density, no vegetation)}: High population and sparse vegetation. An example is San Francisco County, known for its high population density, especially in the city of San Francisco itself. The city's landscape is characterized by relatively sparse vegetation, with a very heavy emphasis on urban development. An aerial view of San Francisco County is presented in Figure E.2 of the Supplementary Material. Figure E.3 illustrates the associated generated scenario.
    
    \item \textbf{Scenario 3 (Medium density intermix)}: Medium population and dense vegetation. An example is parts of Santa Cruz County, particularly around the city of Santa Cruz. Here, there is a medium population density and a mix of urban and rural areas. The county itself is known for its redwood forests and dense vegetation. An aerial view of Santa Cruz County is presented in Figure E.4 of the Supplementary Material. Figure E.5 illustrates the associated generated scenario.
    
    \item \textbf{Scenario 4 (Medium density, no vegetation)}: Medium population and sparse vegetation. An example is Riverside County, which has a medium population density in some areas, particularly in cities like Riverside and Temecula. Most of the county has an arid landscape with little to no vegetation, especially in the desert regions. An aerial view of Riverside County is presented in Figure E.6 of the Supplementary Material. Figure E.7 illustrates the associated generated scenario.
    
    \item \textbf{Scenario 5 (Very low-density vegetation)}: Very low population and dense vegetation. An example is Trinity County. It is sparsely populated and is known for its dense vegetation, including national forests and wilderness areas. An aerial view of Trinity County is presented in Figure E.8 of the Supplementary Materials. Figure E.9 illustrates the associated generated scenario.
    
    \item \textbf{Scenario 6 (Very low density, no vegetation)}: Very low population and sparse vegetation. An example is Modoc County, one of the least densely populated counties in California. It features vast, open spaces with sparse vegetation, and it is located in the northeastern part of the state, known for its high desert and sagebrush landscapes. An aerial view of Modoc County is presented in Figure E.10 of the Supplementary Material. Figure E.11 illustrates the associated generated scenario.
    
\end{itemize}

\subsection{Results of land scenarios}
\textcolor{NavyBlue}{We evaluate all six land scenarios using two fleet sizes (2 and 3 vehicles) and three speed levels (30, 60, and 120 km/h), resulting in 36 model runs. Each scenario is solved with 2 and 3 UAV teams and speeds of 30, 60, or 120 km/h. While current manned firefighting aircraft can operate at higher speeds, our values reflect UAV capabilities and implicitly account for unmodeled delays such as take-off and refilling. Each run is solved using both EM and ERLM. The optimality gap termination criteria is set to $\bar{\epsilon} = 0.03$ for both methods. EM has a fixed time limit of 20 hours. For E-RLM, the total time depends on the number of initial fires, with 3 minutes allocated per fire. This results in 18, 24, and 39 minutes for cases with 4, 6, and 13 fires. We allocate 80 percent of this time to the EM phase and 20 percent to the refill phase (i.e. $r=0.8$). These settings are based on our development experience and work well in all experiments.}

In Section \ref{sec:demo_scenario}, we graphically display the model's output routes within a specific scenario. Then, in Section \ref{sec:land_scenario_summary}, we summarize the results of the six land-based scenarios where we manipulate certain parameters to evaluate their impact on the model's performance.

\subsubsection{Demonstration on a specific scenario}
\label{sec:demo_scenario}
We focus on the high-density intermix scenario, which poses substantial risk in WUI regions due to dense populations and vegetation fueling intense fires. 

\begin{figure}[!ht]
        \centering
        \begin{subfigure}[b]{0.35\textwidth}
            \includegraphics[width=\textwidth]{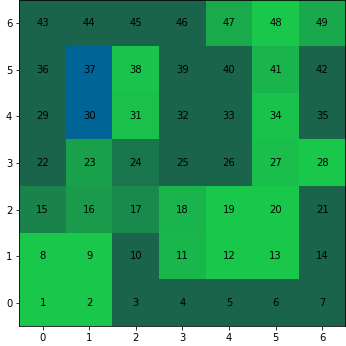}
            \caption{The value of different nodes}
            \label{fig:5_1_value}
        \end{subfigure}
        \quad
        \begin{subfigure}[b]{0.35\textwidth}
            \includegraphics[width=\textwidth]{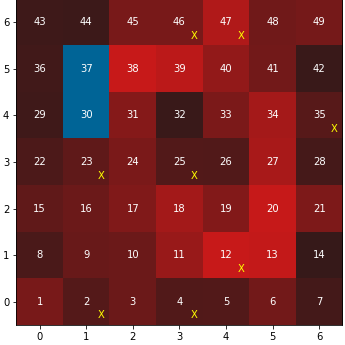}
            \caption{The spread rate of fire at different nodes}
            \label{fig:5_1_spread}
        \end{subfigure}

        \caption{Scenario 1: The high density intermix scenario}
        \label{fig:scenario_1_problem_defnition}

\end{figure}

Figure \ref{fig:scenario_1_problem_defnition}(a) illustrates the default node values (darker green indicates higher value), while Figure \ref{fig:scenario_1_problem_defnition}(b) shows fire spread rates (darker red indicates higher rates). Ignitions are marked with an ``x," and and node 1 is the UAV base. We set the fleet size to 3 (3 UAV teams) and speed to 120 km/h. Node 1 is assumed to be the home base for all three UAVs.

\begin{figure}[!ht]
        \centering
        \begin{subfigure}[b]{0.35\textwidth}
            \includegraphics[width=\textwidth]{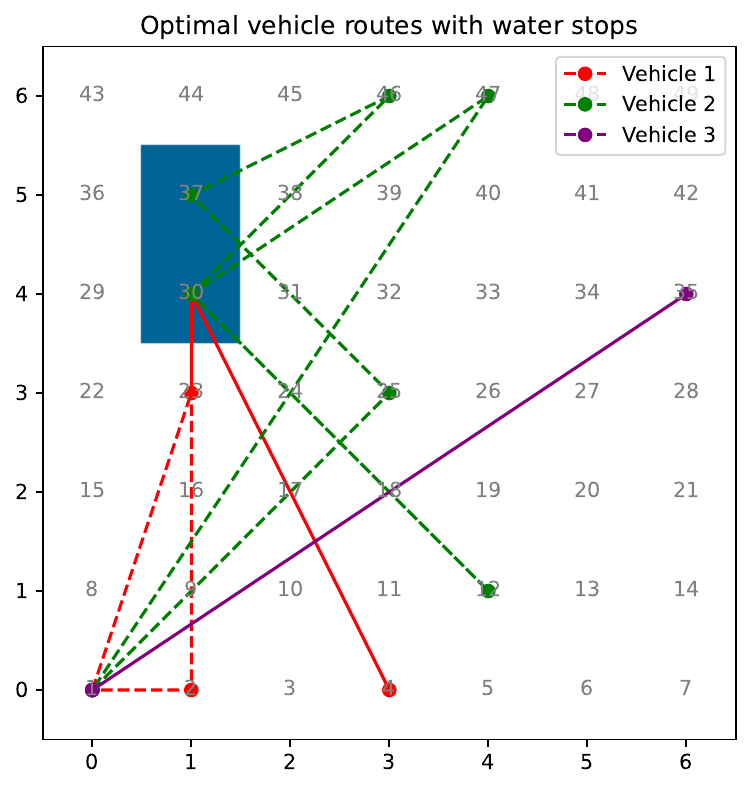}
            \caption{With water stops}
            \label{fig:route_with_water_stops}
        \end{subfigure}
        \quad
        \begin{subfigure}[b]{0.35\textwidth}
            \includegraphics[width=\textwidth]{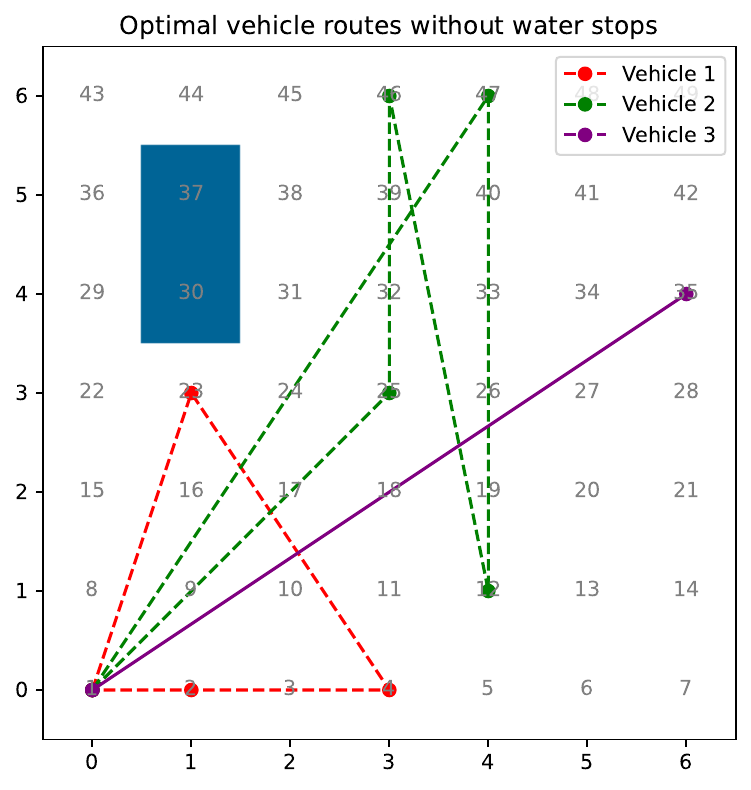}
            \caption{Without water stops}
            \label{fig:route_without_water_stops}
        \end{subfigure}

        \caption{The optimal routes for the 3 UAVs in the scenario shown in Figure \ref{fig:scenario_1_problem_defnition}}

\end{figure}

\textcolor{NavyBlue}{Both EM and E-RLM converge to the same optimal solution with an optimality gap below 1\% in under 10 seconds, successfully extinguishing all eight fires before any spread occurs.} Figures \ref{fig:route_with_water_stops}(a) and \ref{fig:route_without_water_stops}(b) illustrate the optimal routes, with the latter removing water stops for readability.

\subsubsection{Experimenting with a range of land scenarios}
\label{sec:land_scenario_summary}

\textcolor{NavyBlue}{In this section, we report and discuss the results for the six land scenarios described in Section~\ref{subsec:scenarios}. For each scenario, we evaluate six fleet configurations based on two fleet sizes (2 and 3 vehicles) and three speeds (30, 60, and 120~km/h), totaling 36 experimental runs.}

\begin{figure}[ht] 
\centering
\includegraphics[scale=0.95]{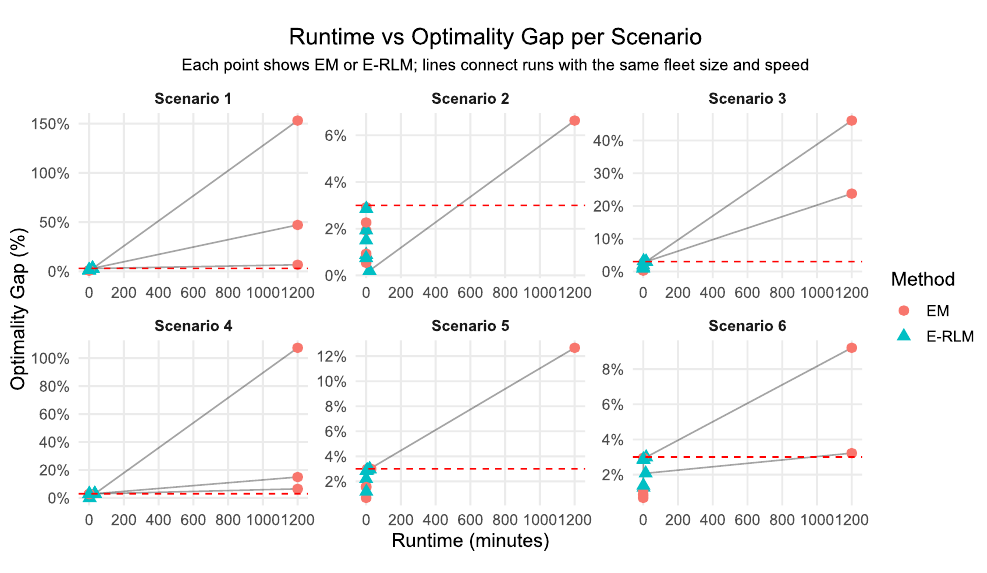}
\caption{\textcolor{NavyBlue}{Comparison of EM and E-RLM in terms of optimality gap and runtime across six scenarios and six fleet configurations.}}
\label{figure:comparison_case_study}
\end{figure}

\textcolor{NavyBlue}{Figure~\ref{figure:comparison_case_study} presents optimality gap (\%) versus runtime (minutes) for EM and E-RLM across all scenarios. Each subplot corresponds to a single scenario and displays 12 data points—six for EM and six for E-RLM—representing the six fleet configurations. EM and E-RLM results are shown as orange-toned circles and blue-toned triangles, respectively. Gray lines connect points from the same configuration across the two methods to highlight their relative performance, and a dashed red line at 3\% marks the optimality gap threshold.}

\textcolor{NavyBlue}{E-RLM consistently outperforms EM in both optimality gap and runtime, particularly under constrained fleet capabilities (e.g., smaller fleet sizes or lower vehicle speeds), across all scenarios. While Figure~\ref{figure:comparison_case_study} illustrates performance trends and the relative advantages of E-RLM, the specific fleet configurations are not explicitly labeled. To supplement this, Table~\ref{table:case_results} presents detailed runtime improvements achieved by E-RLM over EM for each scenario and configuration. Notably, in all scenarios, E-RLM achieves approximately 98\% improvement in configurations with 2 UAVs operating at 30~km/h, where EM struggles to terminate with acceptable performance. Similar substantial gains are observed in other constrained configurations. Also, for settings where EM already performs efficiently—such as with 2 or 3 UAVs at 120~km/h—E-RLM maintains a comparable performance. The average runtime results for E-RLM further underscore its efficiency and scalability: In constrained settings, such as a fleet of 2 UAVs operating at 30 km/h, E-RLM achieves a mean runtime of 21.2 minutes, while increasing the speed to 60 and 120 km/h reduces the average runtime to 10.5 and 0.8 minutes, respectively. Similarly, with 3 UAVs, the mean runtime drops from 17.0 minutes at 30 km/h to just 0.8 and 0.5 minutes at 60 and 120 km/h, respectively. Finally, E-RLM improves the objective function value by more than 0.5\% in 8 out of 36 runs, achieving a mean improvement of 7.85\% with a standard deviation of 12.8\%.}

\begin{table}[ht]
\centering
\color{NavyBlue}
\caption{\textcolor{NavyBlue}{Runtime Improvement (\%) of E-RLM over EM. A value of 0 indicates no improvement (e.g., same performance).}}
\renewcommand{\arraystretch}{1.2}
\begin{tabular}{l ccc ccc}
\toprule
\textbf{Scenario} & \multicolumn{6}{c}{\textbf{Fleet Capability}} \\
 & \multicolumn{3}{c}{2 UAV} & \multicolumn{3}{c}{3 UAV} \\
 & 30 km/h & 60 km/h & 120 km/h & 30 km/h & 60 km/h & 120 km/h \\
\midrule
Scenario 1 & 98 & 98 & 0 & 98 & 0 & 0 \\
Scenario 2 & 98 & 9 & 7 & 0 & 13 & 13 \\
Scenario 3 & 99 & 0 & 0 & 99 & 0 & 0 \\
Scenario 4 & 97 & 97 & 0 & 97 & 0 & 0 \\
Scenario 5 & 98 & 0 & 0 & 29 & 0 & 0 \\
Scenario 6 & 99 & 0 & 0 & 99 & 0 & 0 \\
\bottomrule
\end{tabular}
\label{table:case_results}
\end{table}

\begin{figure}[ht] 
\centering
\includegraphics[scale=1.2]{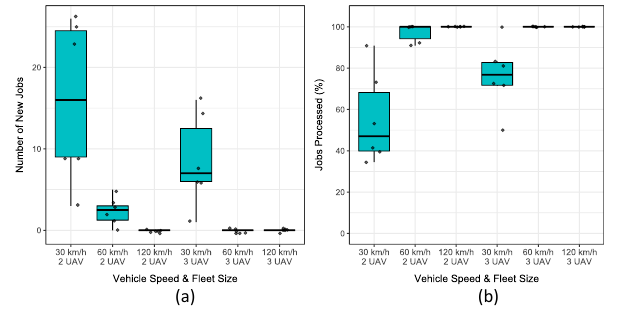}
\caption{\textcolor{NavyBlue}{Impact of fleet configuration on E-RLM performance across all scenarios. (a) Number of new jobs due to fire spread. (b) Percentage of successfully processed jobs.}}
\label{figure:case_study_insights}
\end{figure}

\textcolor{NavyBlue}{We further analyze the solution space across fleet configurations and scenarios in terms of fire spread and process rate. Figure~\ref{figure:case_study_insights} presents key insights from six different settings under optimal E-RLM solutions. Figure~\ref{figure:case_study_insights}(a) shows the number of new jobs (i.e., newly ignited nodes). At 30~km/h with 2 UAVs, fire spread is high and variable (median $\approx$17--18), indicating poor coverage. As speed or fleet size increases, new jobs drop sharply, reaching near zero at 120~km/h regardless of fleet size. Figure~\ref{figure:case_study_insights}(b) illustrates the job processing rate, representing successful fire suppression before node burnout. The lowest performance is at 30~km/h with 2 UAVs (median $\approx$50\%, with some cases below 40\%). In contrast, both 60~km/h and 120~km/h with 2 UAVs achieve approximately 100\% success. Even at 30~km/h, 3 UAVs raise the median rate above 70\%, highlighting the positive impact of fleet size. All settings except ``30~km/h, 2 UAVs'' show strong reliability.}

\textcolor{NavyBlue}{We also investigate the scenario-level impact and find that high vegetation density significantly increases fire spread. Only 5 out of 36 instances resulted in over 10 new jobs (ranging from 14 to 26), all in high-vegetation Scenarios 1, 3, and 5 (see Section~\ref{case_study}). Three involved two UAVs at 30~km/h; two had three UAVs at the same speed. The corresponding job processing rates ranged from 34\% to 73\%, indicating poor coverage. In contrast, higher speed or fleet size consistently pushes processing rates toward 100\%. This shows that while dense vegetation poses a challenge, its impact can be neutralized with sufficient UAV capability. Returning to Figure~\ref{figure:comparison_case_study}, we observe that these same scenarios caused performance issues for the EM model. High vegetation density led to suboptimal solutions and longer run times. This is expected, as faster fire spread generates more jobs, requiring the model to make complex decisions on fire propagation timing and location, as well as more intensive scheduling and routing. This leads to a valuable managerial insight: in areas prone to rapid fire spread, decision-makers should strengthen vehicle availability by increasing fleet size or speed. Further managerial insights are discussed in Section~\ref{sec:managerial_insights}.}

\subsection{Water-rich scenario exploration}
In a real-world context, a ``water" scenario might resemble a collection of islands within a body of water, as illustrated in Figure E.12 in the Supplementary Material. In this scenario, 60\% of the 49 nodes are designated as water resources, which also function as fire-proof, while the remaining nodes are fire-prone. The fire-prone nodes have medium spread rates, with an average spread of 4.4, and a mean node value of 3.3. Four initial fires are started in this scenario.

This scenario can be efficiently managed even with limited vehicle availability due to natural fire breaks and abundant water for firefighting. Table \ref{tab:water-scenario-results} confirms this expectation. New jobs increase only when UAV speed is 30 km/h. \textcolor{NavyBlue}{The E-RLM model consistently achieves optimal solutions with a 100\% job processing rate across all configurations, solving each instance in under 9 minutes.}

\begin{table}[H]
\centering
\caption{E-RLM results for the \textit{Water} scenario across UAV fleet and speed settings}
\label{tab:water-scenario-results}
\begin{tabular}{@{}ccrrrr@{}}
\toprule
\textbf{UAV Fleets} & \textbf{Speed (km/h)} & \textbf{Runtime (s)} & \textbf{Gap (\%)} & \textbf{New Jobs} & \textbf{Jobs Processed (\%)} \\ \midrule
2 & 30  & 520 & 3 & 3 & 100 \\
2 & 60  &  22 & 0 & 0 & 100 \\
2 & 120 &  22 & 2 & 0 & 100 \\
3 & 30  & 496 & 3 & 3 & 100 \\
3 & 60  &  24 & 0 & 0 & 100 \\
3 & 120 &  24 & 2 & 0 & 100 \\
\bottomrule
\end{tabular}
\end{table}

From these observations, we derive two insights. Firstly, the water-based scenario is significantly better in terms of runtime, gap, new jobs created, and percentage of jobs processed. While faster access to water increases the speed with which jobs can be addressed, the water scenario also benefits from water acting as a fire block, as well as there simply being less land to burn. 
Secondly, even in a relatively ideal scenario with ample water access and smaller land areas, limited service availability (2 UAVs at 30 km/h) still affects the efficiency of firefighting UAVs, indicating that reduced vehicle availability constrains wildfire-fighting capabilities even in fire-prone landscapes.


\subsection{Managerial insights}
\label{sec:managerial_insights}

The case study results highlights the significant role of vehicle availability, as determined by fleet size and speed, on both computational performance and operational success. UAV specifications are particularly crucial in scenarios with rapid fire spread and frequent new jobs. In general, larger fleets and faster vehicles are more effective at containing wildfires, reducing new job generation, and extinguishing active fires. \textcolor{NavyBlue}{We test UAV speeds of 30, 60, and 120 km/h to assess their impact. Speed is a tunable parameter that can be adjusted to reflect more precise values in practical applications.}

Fewer jobs are created in scenarios with fewer initial fires and slow fire spread, such as on islands or land surrounded by water, making the use of fewer and slower UAVs viable. While slow UAVs perform adequately in water-based scenarios, medium to fast UAV fleets excel. Notably, UAV fleets that can refill water from nearby sources are especially well-suited for firefighting in water-surrounded areas, such as islands in Hawaii.

Towards the end of a firefighting scenario, fewer jobs are generated as the number of fire-prone nodes decreases. Consequently, a smaller fleet with slower UAVs can manage the remaining tasks, allowing more and faster UAVs to be allocated to areas with early-stage fires that are characterized by higher job creation and rapid spread.

\textcolor{NavyBlue}{Our model is suitable for initial attack planning in high-risk forest areas. The modeled terrain size of 49\,km\textsuperscript{2}  is operationally relevant, capturing critical zones that require immediate intervention. For reference, San Francisco County spans 121\,km\textsuperscript{2}, and more than ten California counties are smaller than 500\,km\textsuperscript{2}. Also, managers may scale the terrain problem to larger areas. Our 1×1 km grid size is defined based on the capacity of a single UAV. If more UAVs are available and allowed to collaborate on the same grid cell, the effective capacity increases, enabling the use of larger cells (e.g., 2×2 km). This allows the total terrain area (e.g., 49 cells × 4\,km\textsuperscript{2} = 196\,km\textsuperscript{2}) to scale without altering the problem structure or increasing computational burden.}

\textcolor{NavyBlue}{Our model adopts a deterministic linear approximation of fire dynamics. Managers should recognize that incorporating uncertainty in fire ignition, spread, and degradation adds realism. To explore this, we conduct a preliminary robustness analysis on our deterministic fire spread modeling by simulating 13 spread rate variations (-30\% to +30\%) across six scenarios and fleet configurations, following the range suggested by Cruz and Alexander (2013). Results show that lower spread rates preserve outcomes, while higher rates can lead to significant variability and losses of up to 85\%. In general, increased fleet capacity mitigates these effects. For example, in Scenario 1 with 2 UAVs at 30 km/h, a +30\% increase reduces the collected value to 15\%, whereas with 3 UAVs at 120 km/h, the loss is only 1\%. These findings motivate a more comprehensive simulation study as a valuable direction for future research.}

\section{Conclusions and future research directions}
We introduce a scheduling and routing challenge centered around the novel concept of degradation-triggered new job arrivals. We study the problem in the context of aerial firefighting. Specifically, we associate each ignited area with a defined action window; delaying intervention results in increased fire size, diminishing the value of the affected area and causing the fire to spread to adjacent regions, thereby necessitating additional action for new ignitions. 

We develop a mixed integer programming model for the solution. Our mathematical model addresses several key decision-making aspects of the problem. These include the initiation of new jobs, specifically the determination of fire spread and the timing and location of new fires. It addresses job selection, vehicle assignment, and the planning of vehicle routes and schedules for the selected jobs. \textcolor{NavyBlue}{We also develop a hybrid solution approach that integrates the mixed-integer programming model into a branch-and-bound framework with dynamic constraint generation. Extensive computational experiments demonstrate the success of the hybrid model, showing its ability to produce high-quality solutions within practical runtime.} To promote the study's reproducibility and further research, we provide open access to our code-base including the mathematical model and a spreadsheet-based user interface. 

Through extensive computational experiments and a case study on California wildfires across seven scenarios with varying population and vegetation densities, we evaluate the model's effectiveness. Fire arrivals are influenced by four adjustable parameters: fleet size, flight speed, initial fire count, and grid spread rates. Results show that larger fleets and faster vehicles are more effective in controlling wildfires, reducing new fires, and extinguishing active ones. In regions prone to rapid fire spread and multiple ignitions, a substantial fleet with advanced capabilities is recommended to manage the increased demand for service.

The critical and complex nature of wildfire management, exacerbated by limited resources, requires sophisticated decision support models. In addition, drawing inspiration from the proverb “a stitch in time saves nine,” our study addresses a relevant problem in which early and well-coordinated intervention can prevent disproportionate damage.

\textcolor{NavyBlue}{Future work could refine the model by incorporating coordination with ground crews, which we omit. We simply fire dynamics, excluding factors such as wind-driven spread and early spotting beyond grid edges. We assume complete suppression after a single intervention, while in practice, multiple actions at the same location may be required. We do not model detailed flight dynamics or operational constraints, such as region-dependent travel times or no-fly zones due to poor visibility. We consider that all fires are triggered internally, without accounting for exogenous fire arrivals. New ignitions caused by uncertain external factors could be incorporated through scenario-based or stochastic extensions. Addressing these aspects could enhance the model’s practical relevance.}

\textcolor{NavyBlue}{We acknowledge that the performance of the hybrid model may depend on the quality of the initial solution from the exact model, due to the risk of prematurely pruning viable branches in the relaxed model (see Section 4.2.1). While we do not observe this issue in our experiments, future work may explore scalability on larger instances. In such cases, tuning and allocating more time to the exact model could help ensure better initial solutions.}

Future research may also explore the complexities of multiple objectives inherently encompassed by our problem. Specifically, these objectives may include prioritizing the protection of valuable areas, reducing the time needed to control fires, and determining the most effective fleet size.

\section*{Acknowledgement}
We are grateful to the University at Buffalo for allowing us access to their Center for Computational Research for high-performance computing resources. This work was conducted during the research visit of the first author at the University at Buffalo, funded by the TUBITAK 2219 scholarship. \textcolor{NavyBlue}{We are also grateful to the anonymous reviewers for their constructive comments and recommendations, which greatly helped improve the quality and clarity of the manuscript}.

\bibliography{mybibfile}

\end{document}